\documentclass{article}

\usepackage[nonatbib,preprint]{style}
\usepackage[page,header]{appendix}
\usepackage{titletoc}
\usepackage{setspace}
\usepackage[utf8]{inputenc} 
\usepackage[T1]{fontenc}    
\usepackage{url}            
\usepackage{booktabs}       
\usepackage{amsfonts}       
\usepackage{nicefrac}       
\usepackage{microtype}      

\usepackage{amsmath,amssymb,amsthm}
\usepackage[ruled,vlined]{algorithm2e}
\usepackage{algpseudocode}
\usepackage{graphicx}
\usepackage{float}
\usepackage{wrapfig}
\usepackage{subcaption}
\usepackage{caption}
\usepackage{mathtools}
\usepackage{xcolor}
\usepackage{verbatim}
\usepackage{cancel}
\usepackage[user,titleref]{zref}
\usepackage{cancel}
\usepackage{hyperref}  
\usepackage{enumitem}  
\usepackage{chngcntr}  

\definecolor{darkgreen}{RGB}{40, 150, 40}
\definecolor{gray}{RGB}{140, 140, 140}
\definecolor{darkmagenta}{HTML}{9c027b}
\newcommand{\rulesep}{\color{gray}\unskip\ \vrule\ }

\DeclareMathOperator*{\krdim}{krdim}
\DeclareMathOperator*{\codim}{codim}

\newcommand{\btau}{{\pmb{\tau}}}

\newcommand{\R}{{\mathbb{R}}}
\newcommand{\C}{{\mathbb{C}}}
\newcommand{\E}{{\mathbb{E}}}
\newcommand{\M}{{\mathcal{M}}}

\newcommand{\OO}{{\mathcal{O}}}

\newcommand{\mf}{{\mathfrak{f}}}
\newcommand{\rar}{\rightarrow}
\newcommand{\lar}{\leftarrow}

\newcommand{\tpl}{T\text{+}L}
\newcommand{\norm}[1]{\left\lVert#1\right\rVert}

\newtheorem{definition}{Definition}[section]
\newtheorem{theorem}{Theorem}[section]

\newtheorem{lemma}{Lemma}[section]

\newtheoremstyle{remark}
  {}   
  {}   
  {\upshape}  
  {}       
  {\bfseries} 
  {.}         
  {} 
  {} 

\newtheorem*{definition31}{Definition 3.1}
\newtheorem*{definition32}{Definition 3.2}
\newtheorem*{theorem31}{Theorem 3.1}
\newtheorem*{definition33}{Definition 3.3}
\newtheorem*{theorem32}{Theorem 3.2}
\newtheorem*{lemma31}{Lemma 3.1}
\newtheorem*{definition34}{Definition 3.4}
\newtheorem*{lemma32}{Lemma 3.2}

\makeatletter
\g@addto@macro\normalsize{%
  \setlength\abovedisplayskip{6pt}
  \setlength\belowdisplayskip{6pt}
  \setlength\abovedisplayshortskip{6pt}
  \setlength\belowdisplayshortskip{6pt}
  \setlength{\jot}{0pt}
}

\newlength{\mylen}
\setbox1=\hbox{$\bullet$}\setbox2=\hbox{\tiny$\bullet$}
\setlist[itemize]{itemsep=0mm, topsep=2pt}

\usepackage{etoolbox}

\SetCommentSty{mycommfont}

\makeatletter
\patchcmd{\@algocf@start}
  {-1.5em}
  {0pt}
  {}{}
\makeatother

\makeatletter
\newcommand\fs@nobottomruled{\def\@fs@cfont{\bfseries}\let\@fs@capt\floatc@ruled
  \def\@fs@pre{\hrule height.8pt depth0pt \kern2pt}%
  \def\@fs@post{}
  \def\@fs@mid{\kern2pt\hrule\kern2pt}%
  \let\@fs@iftopcapt\iftrue}
\makeatother

\title{Analytic Manifold Learning: Unifying \& Evaluating Representations for Continuous Control}
\newcommand{\ouralgo}{AML}
\newcommand{\ouralgofullname}{Analytic Manifold Learning}
\author{%
  Rika Antonova \\
  EECS, KTH, Stockholm, Sweden \\
  \texttt{rika.antonova@gmail.com} \\
  \And
  Maksim Maydanskiy \\
  \texttt{maksim.m@gmail.com} \\
  \AND
  Danica Kragic \\
  EECS, KTH, Stockholm, Sweden \\
  \And
  Sam Devlin \\
  Microsoft Research \\
  \And
  Katja Hofmann \\
  Microsoft Research \\
}
\begin{document}

\maketitle

\begin{abstract}
We address the problem of learning reusable state representations from streaming high-dimensional observations. This is important for areas like Reinforcement Learning (RL), which yields non-stationary data distributions during training. We make two key contributions. First, we propose an evaluation suite that measures alignment between latent and true low-dimensional states. We benchmark several widely used unsupervised learning approaches. This uncovers the strengths and limitations of existing approaches that impose additional constraints/objectives on the latent space.
Our second contribution is a unifying mathematical formulation for learning latent relations. We learn analytic relations on source domains, then use these relations to help structure the latent space when learning on target domains. This formulation enables a more general, flexible and principled way of shaping the latent space. It formalizes the notion of learning independent relations, without imposing restrictive simplifying assumptions or requiring domain-specific information. We present mathematical properties, concrete algorithms for implementation and experimental validation of successful learning and transfer of latent relations.
\end{abstract}

\section{Introduction}

In this work, we address the problem of learning reusable state representations from streaming high-dimensional observations. Consider the case when a deep reinforcement learning (RL) algorithm is trained on a set of source domains. Low-dimensional state representations could be extracted from intermediate layers of RL networks, but they might not be reusable on a target domain with different rewards or dynamics. To aid transfer and ensure non-degenerate embeddings, it is common to add unsupervised learning objectives. However, the quality of resulting representations is usually not evaluated rigorously. Moreover, constructing and prioritizing such objectives is done manually: auxiliary losses are picked heuristically and hand-tuned for transfer to a new set of domains or tasks. 

As the first part of our contribution, we provide a set of tools and environments to improve evaluation of learning representations for use in continuous control. We evaluate commonly used unsupervised approaches and explain new insights that highlight the need for critical analysis of existing approaches.
Our evaluation suite provides tools to measure alignment between the latent state from unsupervised learners and the true low-dimensional state from the physics simulator. 
Furthermore, we introduce new environments for manipulation with multiple objects and ability to vary their complexity: from geometric shapes to mesh scans and visualizations of real objects.
We show that, while alignment with true state is achieved on the simpler benchmarks, new environments present a formidable challenge: existing unsupervised objectives do not guarantee robust and transferable state representation learning. 

The second part of our contribution is a formalization of learning latent objectives from a set of source domains. 
We describe the mathematical perspective of this approach as finding a set of functionally independent relations that hold for the data sub-manifold. We explain theoretical properties and guarantees that this perspective offers. 
Previous work constructed latent relations based on domain knowledge or algorithmic insights, e.g. using continuity~\cite{wiskott2002slow}, mutual information with prior states~\cite{anand2019unsupervised}, consistency with a forward or inverse model (see~\cite{lesort2018state} for a survey).
Our formulation offers a unified view, allowing to leverage known relations, discover new ones and incorporate relations into joint training for transfer to target domains.
We describe algorithms for concrete implementation and visualize the learned relations on analytic and physics-based domains. In our final set of experiments, we show successful transfer of relations learned from source domains with simple geometric shapes to target domains that contain objects with real textures and 3D scanned meshes. We also show that our approach obtains improved latent space encoder mappings with smaller distortion variability.

\section{Evaluation Suite for Unsupervised Learning for Continuous Control}
\label{sec:suite}

\begin{figure}[t]
\begin{subfigure}{.067\textwidth}
\includegraphics[width=1.0\textwidth]{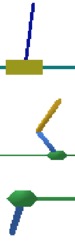}
\end{subfigure}
\begin{subfigure}{.097\textwidth}
\includegraphics[width=1.0\textwidth]{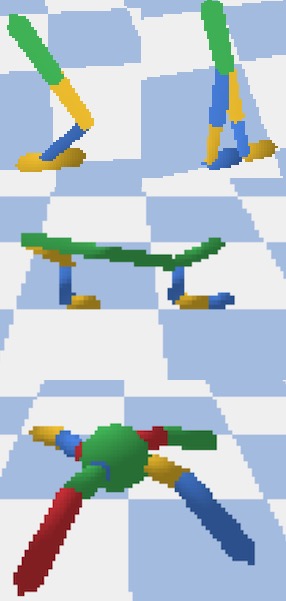}
\end{subfigure}
\begin{subfigure}{.14\textwidth}
\centering
\includegraphics[width=1.0\textwidth]{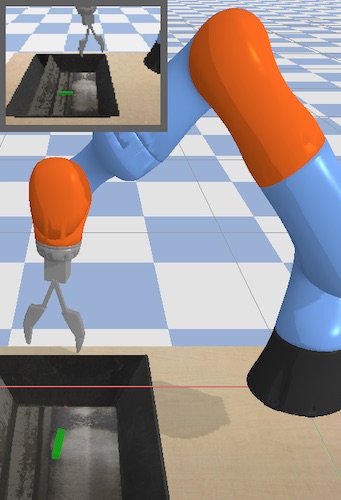}
\end{subfigure}
\vrule width 2pt
\vspace{1px}
\begin{subfigure}{.56\textwidth}
\includegraphics[width=1.0\textwidth]{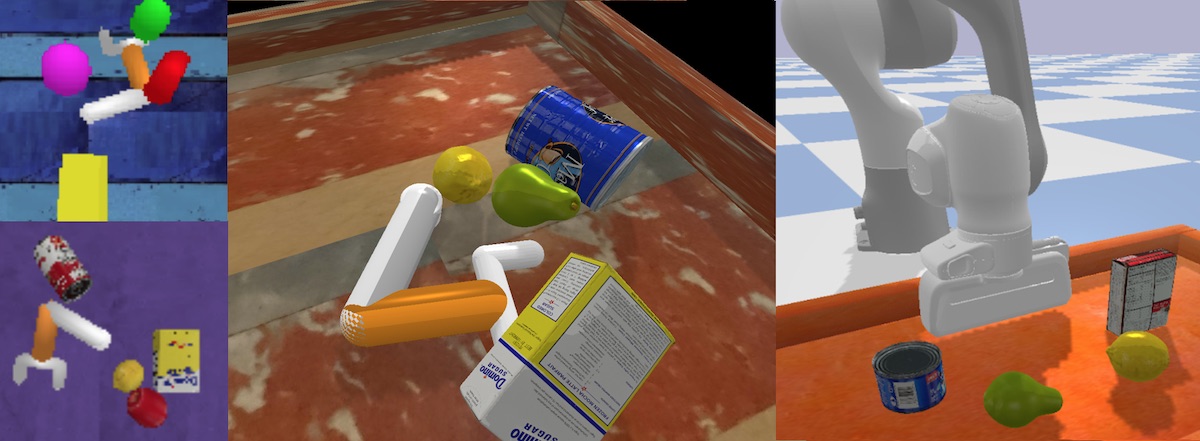}
\end{subfigure}
\begin{subfigure}{.10\textwidth}
\includegraphics[width=1.0\textwidth]{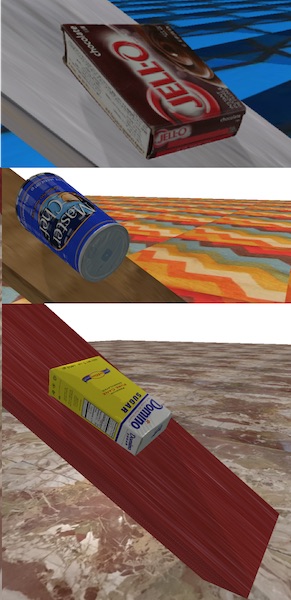}
\end{subfigure}
\vspace{-7px}
\caption{Evaluation suite environments. Left: Standard PyBullet envs for which our suite yields both pixels and low-dimensional state.
Right: Proposed new advanced domains with YCB objects.}
\label{fig:envs}
\vspace{-13px}
\end{figure}

Reinforcement learning (RL) has shown strong progress  recently~\cite{franccois2018introduction}, and
RL for continuous control is particularly promising for robotics~\cite{kober2013reinforcement}. However, training for each robotics task from scratch is prohibitively expensive,
especially for high-dimensional observations.
Unsupervised learning could help obtaining low-dimensional latent representations, e.g. with variational autoencoder (VAE)~\cite{kingma2013auto} variants.
However, evaluation of these mostly focused on datasets, with a limiting assumption that the training data distribution is stationary~\cite{deeprobo2018limits}. Moreover, advanced approaches usually report best-case results, achieved only with exact parameters that the authors find to work for a given static dataset. Obtaining reconstructions that are clear enough to judge whether all important information is encoded in the latent state could still require days or weeks of training~\cite{IODINE19, heljakka2018pioneer}.
These limitations severely impair the adoption of unsupervised representation learning in robotics. In stark contrast to the learning community, a vast majority of roboticists still need to rely on hand-crafted low-dimensional features.

We propose an evaluation suite that helps analyze the alignment between the learned latent state and true low-dimensional state. Unsupervised approaches receive frames that an RL policy yields during its own training: a non-stationary stream of RGB images.
The alignment of the learned latent state and the true state is measured periodically as training proceeds. For this, we do a regression fit using a small fully-connected neural network, which takes latents as inputs and is trained to produce low-dimensional states as outputs (position \& orientation of objects; robot joint angles, velocities, contacts). The quality of alignment is characterized by the resulting test error rate. This approach helps quantify latent space quality without the need for detailed reconstructions. 
To connect our suite to existing benchmarks, we extend the OpenAI gym interface~\cite{brockman2016openai} of widely used robotics domains so that both pixel- and low-dimensional state is reported.
We use an open source simulator: PyBullet~\cite{coumans2019}. Simulation environments are parallelized, ensuring scalability.
We introduce advanced domains utilizing meshes from 3D scans of real objects from the YCB dataset~\cite{calli2015ycb}.
This yields realistic object appearances 
and dynamics.
Our \textit{RearrangeYCB} domain models object rearrangement tasks, with variants for using realistic vs basic robot arms. The \textit{RearrangeGeom} domain offers an option with simple geometric shapes instead of object scans. The \textit{YCB-on-incline} domain models objects sliding down an incline, with options to change friction and apply external forces; \textit{Geom-on-incline} offers a variant with simple single-color geometric shapes.
Figure~\ref{fig:envs} gives an overview.

\subsection{Benchmarking Latent State Alignment of Unsupervised Approaches}
\label{sec:bench}

\begin{figure}[t]
\centering
\begin{subfigure}{.235\textwidth}
\centering
\includegraphics[width=1\textwidth]{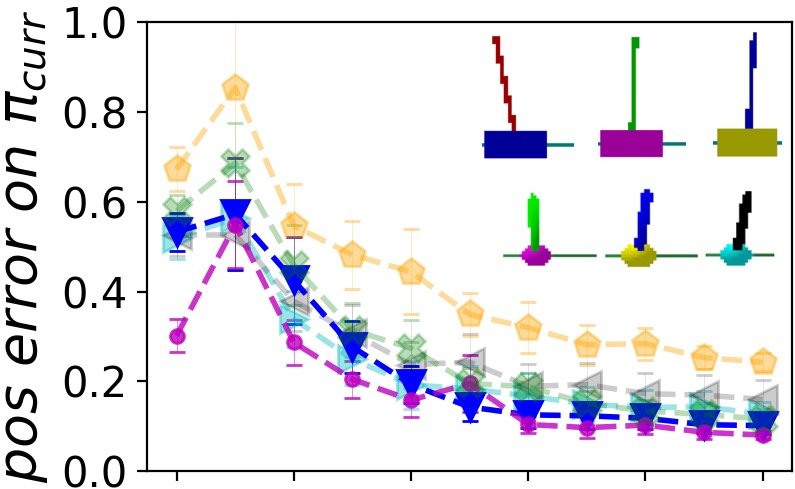}
\includegraphics[width=1\textwidth]{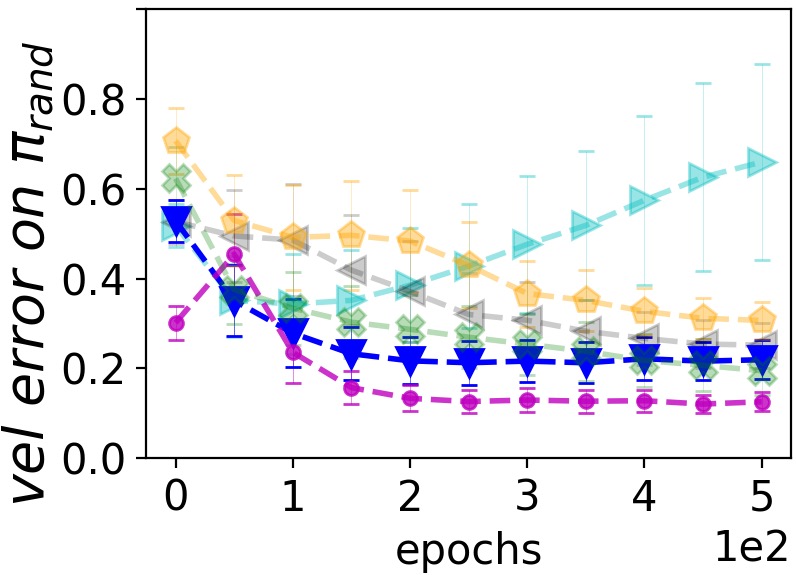}
\end{subfigure}
\begin{subfigure}{.235\textwidth}
\centering
\includegraphics[width=1\textwidth]{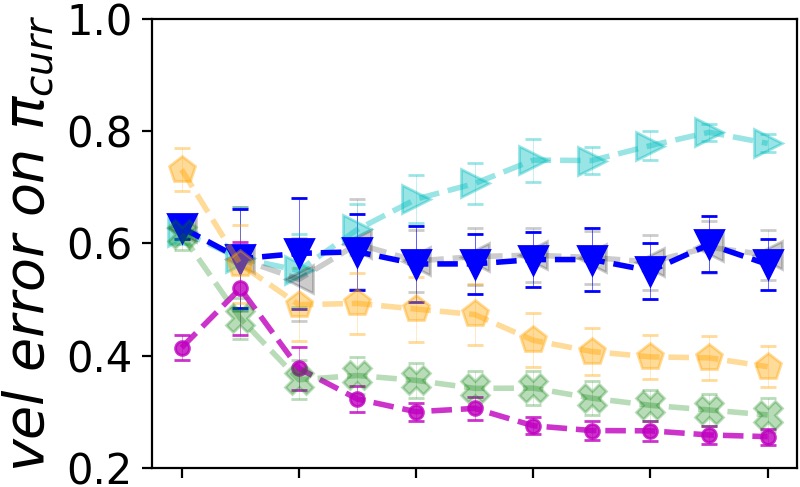}
\includegraphics[width=1\textwidth]{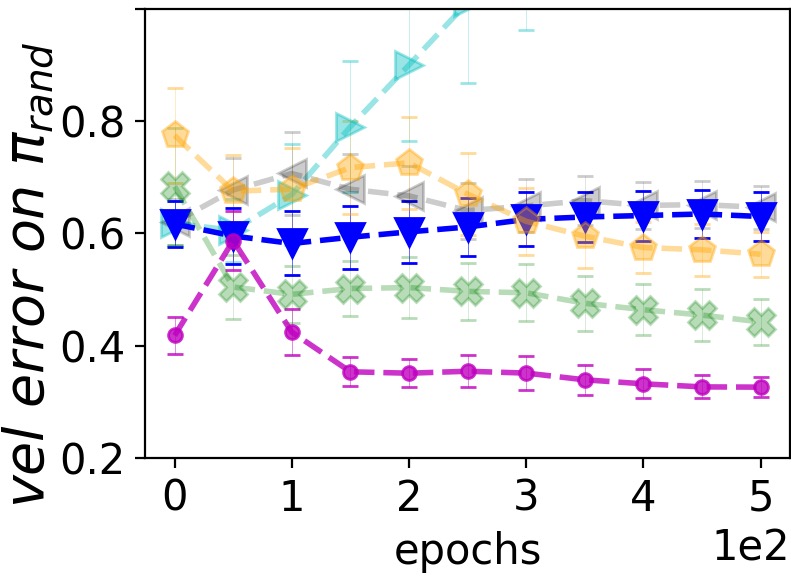}
\end{subfigure}
\rulesep
\begin{subfigure}{.245\textwidth}
\centering
\includegraphics[width=1.0\textwidth]{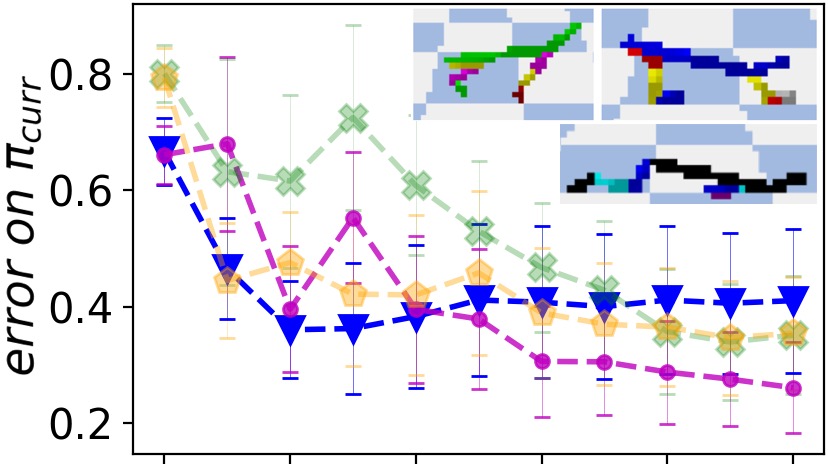}
\includegraphics[width=1.0\textwidth]{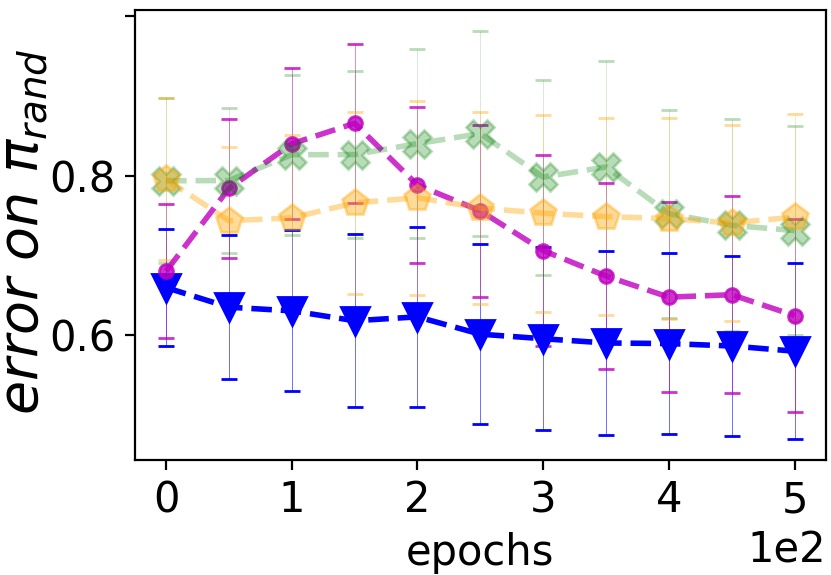}
\end{subfigure}
\rulesep
\begin{subfigure}{.235\textwidth}
\centering
\includegraphics[width=1\textwidth]{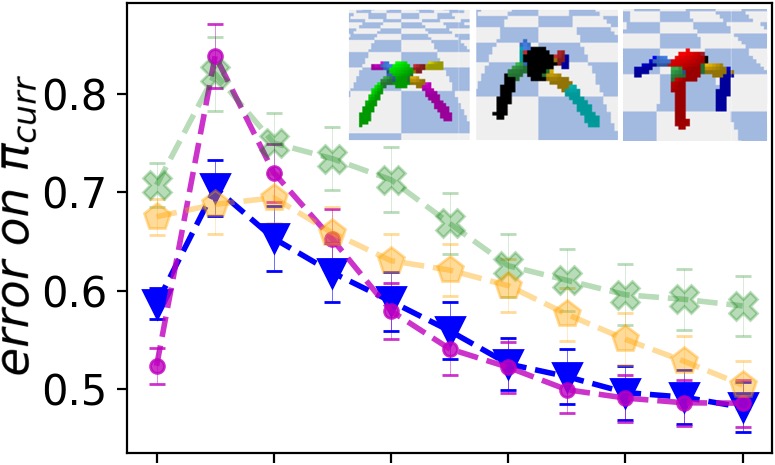}
\includegraphics[width=1\textwidth]{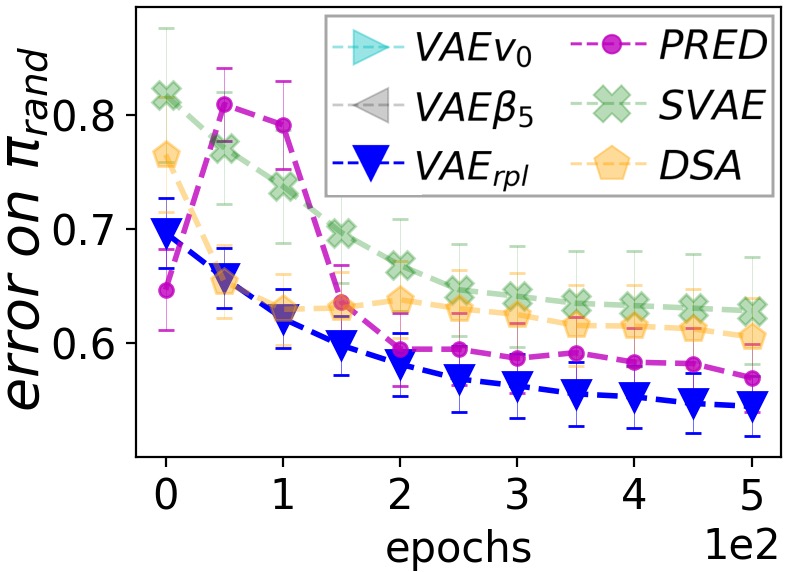}
\end{subfigure}
\caption{Benchmarking alignment with true low-dimensional state. Plots show mean test error of NN regressors trained with current latent codes as inputs and true states (robot positions, velocities, contacts) as outputs. 90\% confidence intervals over 6 training runs for each unsupervised approach are shown ($>\!\!140$ training runs overall). Training uses frames from replay buffers (1024 frames per batch; 10 batches per epoch, 50 for locomotion). Top row: performance on frames from current RL policy $\pi_{curr}$, middle row: random policy $\pi_{rand}$. 
1st column shows results for \textit{CartPole} and \textit{InvertedPendulum} for position \& angle; 2nd column: for velocity. 3rd column shows aggregated results for position, velocity and contacts for \textit{HalfCheetah}; 4th column shows these for \textit{Ant} domain.}
\label{fig:bench}
\vspace{-15px}
\end{figure}

\begin{figure}[b]
\centering
\vspace{-12px}
\begin{subfigure}{.44\textwidth}
\centering
\includegraphics[width=1\textwidth]{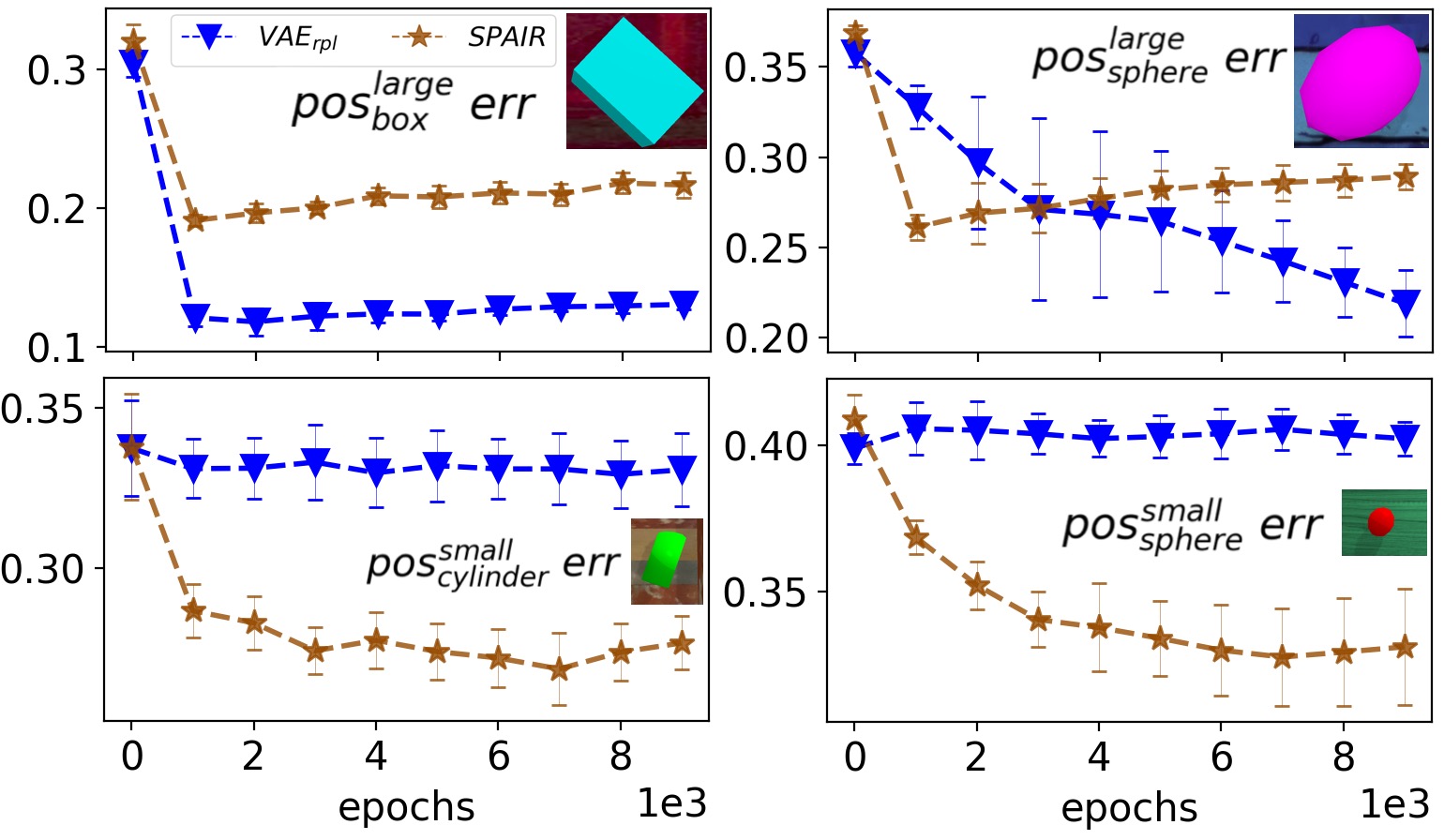}
\end{subfigure}
\begin{subfigure}{.44\textwidth}
\centering
\includegraphics[width=1\textwidth]{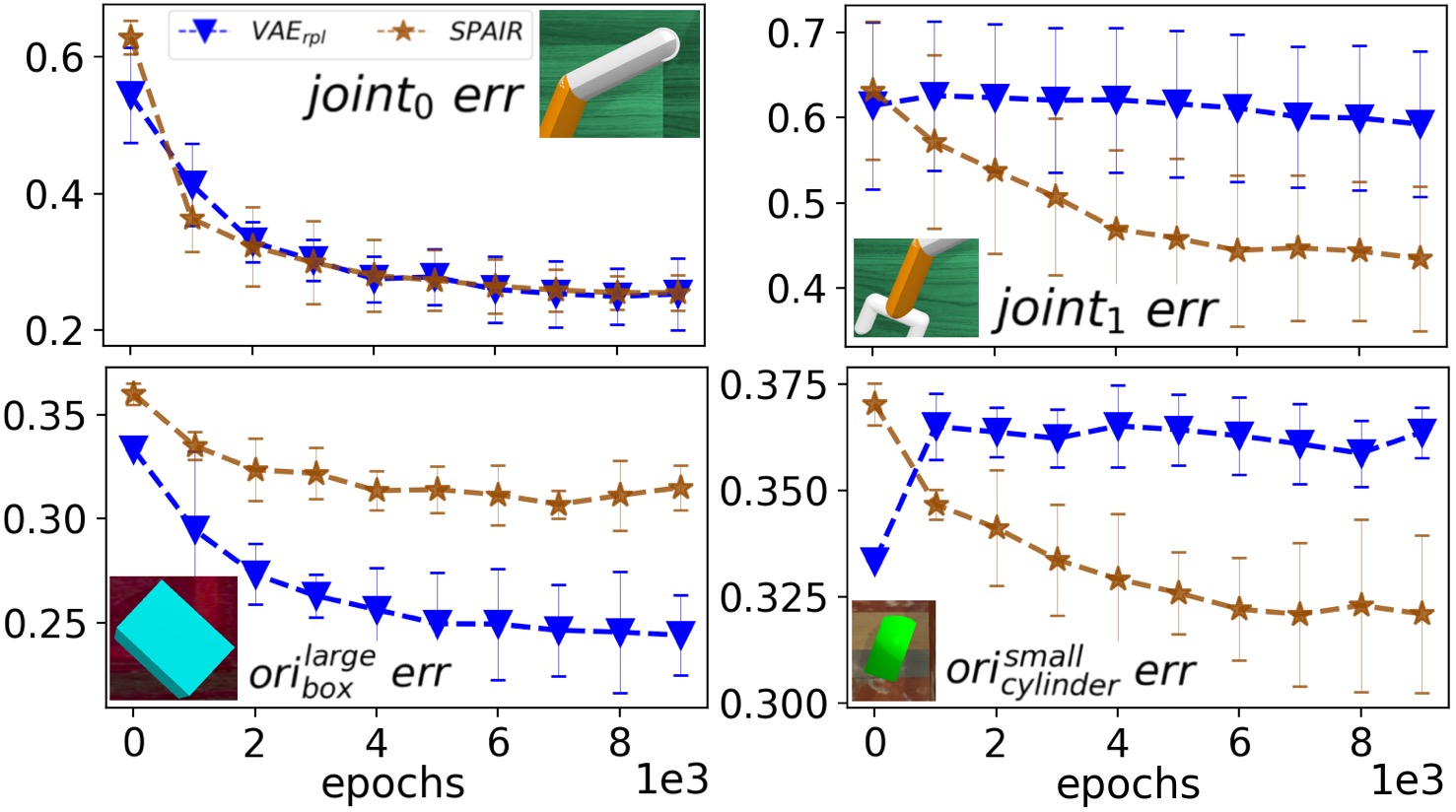}
\end{subfigure}
\begin{subfigure}{.1\textwidth}
\centering
\includegraphics[width=1\textwidth]{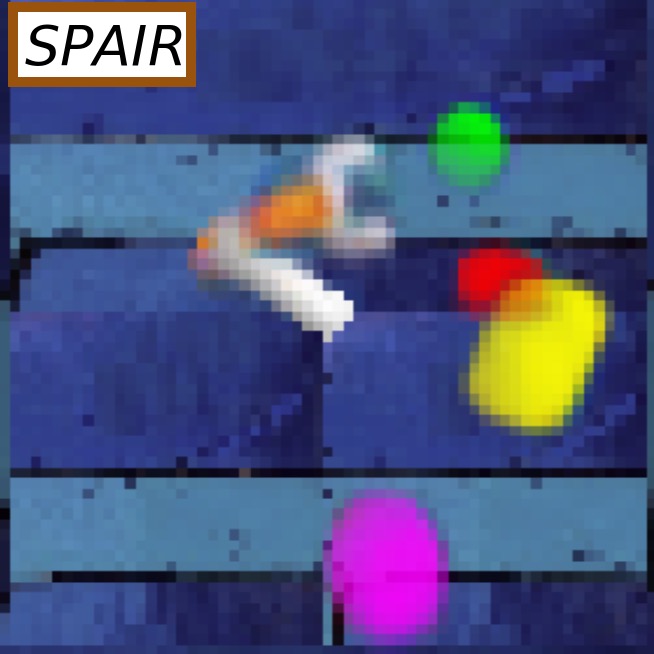}
\vspace{-8px}
\includegraphics[width=1\textwidth]{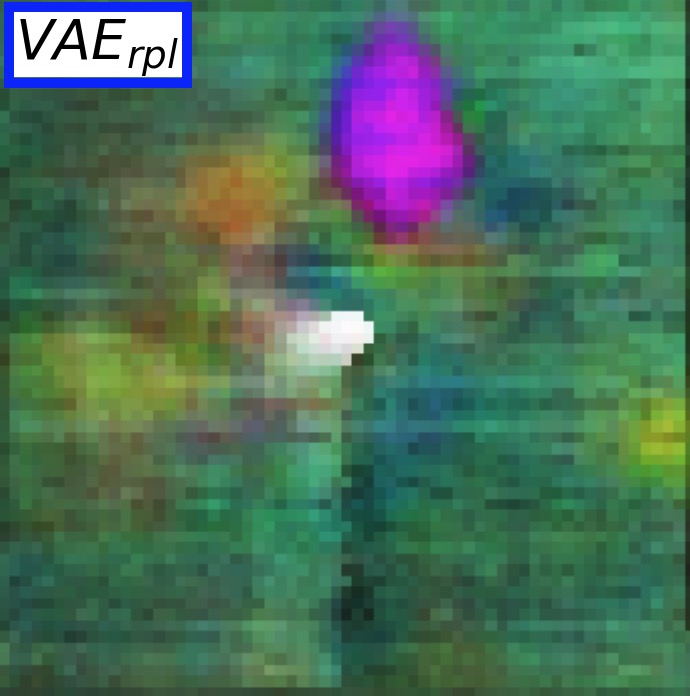}
\vspace{5px}
\end{subfigure}
\vspace{-5px}
\caption{Evaluation on the \textit{RearrangeGeom} domain (reconstructing YCB objects was difficult for existing approaches, so \textit{RearrangeYCB} was too challenging). $V\!\!AE_{rpl}$ encoded angle of the main robot joint, location \& partly orientation (major axis) of the largest objects. $SP\!AIR$ encoded (rough) locations quickly, but did not improve with longer training (bounding boxes not tight).}
\label{fig:spair}
\vspace{-1px}
\end{figure}

To demonstrate usage and benefits of the suite we evaluated several widely used and recently proposed unsupervised learning approaches.
Unsupervised approaches get 64x64 pixel images sampled from replay buffers, filled by PPO RL learners~\cite{schulman2017proximal}.
Figures~\ref{fig:bench},~\ref{fig:spair} show comparisons for several commonly used unsupervised approaches (Appendix~\ref{sec:suppl_a} gives more detailed descriptions and learning parameters).
\vspace{1px}
$\pmb{V\!\!AE_{v_0}}$~\cite{kingma2013auto}: a VAE with a 4-layer convolutional encoder and corresponding de-convolutional decoder;
$\pmb{V\!\!AE_{rpl}}$: a VAE with a replay buffer that retains 50\% of frames from beginning of training (our modification of VAE for improved performance on a wider range of RL policies);
$\pmb{\beta\text{-}V\!\!AE}$~\cite{higgins2017beta}: a VAE with $\beta$ parameter to encourage disentanglement (we tried several $\beta$ parameters and also included the replay enhancement from $V\!\!AE_{rpl}$);
$\pmb{SV\!\!AE}$: a sequential VAE that reconstructs a sequence of frames $x_{1},...,x_{T}$;
$\pmb{P\!RED}$: a VAE that, given a sequence of frames  $x_{1},...,x_{T}$, constructs a predictive sequence $x_{1},...,x_{T\text{+}L}$;
$\pmb{DS\!A}$~\cite{yingzhen2018disentangled}: a sequential autoencoder that uses structured variational inference to encourage separation of static and dynamic aspects of the latent state;
$\pmb{SP\!AIR}$~\cite{SPAIR19}: a spatially invariant and faster version of AIR~\cite{AIR16} that imposes a particular structure on the latent state.

Figure~\ref{fig:bench} shows results on multicolor versions of \textit{CartPole}, \textit{InvertedPendulum}, \textit{HalfCheetah} and \textit{Ant} domains (multicolor to avoid learning trivial color-based features).
We evaluated using two kinds of policies: a current RL learner policy $\pi_{curr}$, and a random policy $\pi_{rand}$.
Success on $\pi_{rand}$ is needed for transfer: when learning a new task, initial frames are more similar to those from a random policy than a final source task policy. $V\!\!AE_{v_0}$ performed poorly on $\pi_{rand}$. We discovered that this can be alleviated by replaying frames from initial random policy. The resulting $V\!\!AE_{rpl}$ offers good alignment for positions.
Surprisingly, $\beta\text{-}V\!\!AE$ offered no improvement over $V\!\!AE_{rpl}$. We used $\beta \!\in\! \{100,20,10,5,0.5\}$; the best ($\beta\!=\!5$) performed slightly worse than $V\!\!AE_{rpl}$ on pendulum domains (shown in Figure~\ref{fig:bench}), the rest did significantly worse (omitted from plots).
Sequential approaches $SV\!\!AE$,$P\!RED$,$DS\!A$ offered significant gains when measuring alignment for velocity. Despite its simpler architecture, $P\!RED$ performed best on pendulum domains. For aggregated performance on position, velocity and contacts (i.e. whether robot joints touch the ground) for locomotion: PRED outperformed $V\!\!AE_{rpl}$ on $\pi_{curr}$, but was second-best on $\pi_{rand}$. Overall, this set of experiments was illuminating: simpler approaches were often better than more advanced ones.

For our newly proposed domains with multiple objects: the first surprising result was that all of the approaches we tested failed to achieve clear reconstructions of objects from the YCB dataset. This was despite attempts to use larger architectures, up to 8 layers with skip connections, similar to~\cite{park2019deepsdf}.
Figure~\ref{fig:spair} shows results for $SP\!AIR$ vs $V\!\!AE_{rpl}$. $SP\!AIR$ succeeded to reconstruct \textit{RearrangeGeom}, while other approaches failed. This indicates that our multi-object benchmark is a highly needed addition to the current continuous control benchmark pool. While single-object benchmarks might still be challenging for control, they could be inherently simpler for latent state learning and reconstruction.

Overall, our analysis shows that structuring the latent space can be beneficial, but has to be done such that it does not impair the learning process and resulting representations. This is not trivial, since seemingly beneficial objectives that worked well in the past could be detrimental on new domains. However, forgoing structure completely can fail on more advanced scenes. Hence, in the following sections we show an alternative direction: a principled way to learn a set of general rules from source domains, then apply them to structure latent space of unsupervised learners on target domains.

\section{{\ouralgofullname}}

We now motivate the need to unify learning latent relations, then provide a rigorous and general mathematical formulation for this problem.
Let $x_t$ denote a high-dimensional (observable) state at time $t$ and $s_t$ denote the corresponding low-dimensional or latent state. $x_t$ could be an RGB image of a scene with a robot \& objects, while $s_t$ could contain robot joint angles, object poses, and velocities.
Consider an example of a latent relation: the continuity (slowness) principle~\cite{wiskott2002slow, kompella2011incremental}. It postulates continuity in the latent states, implying that sudden changes are unlikely. It imposes a loss $L_{cont}(\mathcal{D}_x, \phi) = \E\big[ || s_{t+1} - s_{t} ||^2\big]$, with $D_x\!\!=\!\!\{x_t,x_{t+1}, ...\}$ and encoder $\phi(x) \!=\! s$. A related heuristic from~\cite{anand2019unsupervised} maximizes mutual information between parts of consecutive latent states.
Such approaches may be viewed as postulating concrete latent relations: $g (s_t,s_{t+1})= c_\epsilon$, where $g$ is the squared distance between $s_t$ and $s_{t+1}$ for $L_{cont}$, and a more complicated relation for ~\cite{anand2019unsupervised}.
Ultimately, all these are heuristics coming from intuition or prior knowledge. However, only a subset of them might hold for a given class of domains. Moreover, it would be tedious and error-prone to manually compose and incorporate a comprehensive set of such heuristics into the overall optimization process.

We take a broader perspective. Let $g(\mathcal{D}_{\tau})\!=\!0$ define a relation that holds on a set of sequences $\mathcal{D}_{\tau}\!=\!\{ \tau^{(i)} \}_{i=1}^M$.
$\mathcal{D}_{\tau}$ could contain state sequences $\tau \!=\! [s_t, ..., s_{t+T}]$ from a set of source domains.
We start by learning a relation $g_1$; then learn $g_2$ that differs from $g_1$; then learn $g_3$ different from $\{g_1,g_2\}$ and so on. 
Overall, we aim to learn a set of relations that are (approximately) independent, and we define independence rigorously.
To understand why rigor is important here, recall the significance of the definition of independence in linear algebra: it is central to the theory and algorithms in that field. Extending the notion of independence to our more general nonlinear setting is not trivial, since naive definitions can yield unusable results. Our contribution is developing rigorous definitions of independence, and ensuring the result can be analyzed theoretically \& used for practical algorithms.

\subsection{Mathematical Formulation}
\label{sec:math}

Let $\R^N$ be the ambient space of all possible latent state sequences $\tau$ (of some fixed length). Let $\M$ be the submanifold of actual state sequences that a dynamical system from one of our domains could generate (under any control policy).
A common view of discovering $\M$ is to learn a mapping that produces only plausible sequences as output (the `mapping' view).
Alternatively, a submanifold  can be specified by describing all equations (i.e. relations) that have to hold for points in the submanifold. 

We are interested in finding relations that are in some sense independent. In linear algebra, a dependency is a linear combination of vectors with constant coefficients. In our nonlinear setting the analogous notion is that of \textit{syzygy}.
A collection of functions $\mf^\ddagger \!=\!\{f_1,...,f_k\}$ is called a \textit{syzygy} if $\sum_{j=0}^{k} f_j g_j$ is zero. Observe that this sum is a linear combination of relations $g_1,...,g_k$ with coefficients in the ring of functions.
If there is no syzygy $\mf^\ddagger$ s.t. $\sum_{j=0}^{k} f_j g_j \!=\! 0$, then $g_1,...,g_k$ are independent.
However, this notion of independence is too general for our case, since it deems any $g_1,g_2$ dependent: $g_1 \cdot g_2 - g_2 \cdot g_1=0$ holds for any $g_1,g_2$.
Hence, we define \textit{restricted syzygies}.

\vspace{4px}
\begin{definition}[Restricted Syzygy]
\label{def:restricted_syzygy}
Restricted syzygy for relations $g_1,...,g_k$ is a syzygy  with the last entry $f_k$ equal to $-1$, i.e.  
$\mf = \{f_1,..., f_{k-1},f_k\!=\!-1\}$ with $\sum_{j=1}^k f_j g_j \!=\! 0.$
\end{definition}
\vspace{2px}
\begin{definition}[Restricted Independence]
\label{def:restricted_indep}
$g_k$ is independent from $g_1,...,g_{k-1}$ in a restricted sense if the equality $\sum_{j=1}^k f_j g_j \!=\! 0$ implies $f_k\neq-1$, i.e. if there exists no restricted syzygy for $g_1,...,g_k$.
\end{definition}
\vspace{-4px}
For $\mf\!=\!\{f_1,..., f_{k-1}, f_k\!=\!-1\}$ we denote $\sum_{j=1}^k f_j(\tau) g_j(\tau)$ by $\mf(\tau, g_1, ..., g_k)$.
Using the above definitions, we construct a practical algorithm (Section~\ref{sec:aml_algo}) for learning independent relations.
The overall idea is: while learning $g_k$s, we are also looking for restricted syzygies $\mf(\tau, g_1, ..., g_k) \!=\! 0$. Finding them would mean $g_k$s are dependent, so we augment the loss for learning $g_k$ to push it away from being dependent. We proceed sequentially: first learning $g_1$, then $g_2$ while ensuring no restricted syzygies appear for $\{g_1,g_2\}$, then learning $g_3$ and so on. Section~\ref{seq:related_work} explains motivations for learning sequentially.
For training $g_k$s we use \textit{on-manifold} data: $\tau$ sequences from our dynamical system.
Restricted syzygies $\mf$ are trained using \mbox{\textit{off-manifold}} data: $\tau_{o\!f\!f} \!=\! \{s_{o\!f\!f_t},s_{o\!f\!f_{t+1}},...,s_{o\!f\!f_T}\}$, because we aim for independence of $g_k$s on $\R^N$, not restricted to $\M$ (on $\M$ $g_k$s should be zero). $\tau_{o\!f\!f}$ do not lie on our data submanifold and can come from thickening of on-manifold data or can be random (when $\R^N$ is large, the probability a random sequence satisfies equations of motion is insignificant).
Independence in the sense of Definition~\ref{def:restricted_indep} is the same as saying that $g_k$ does not lie in the \textit{ideal} generated by $(g_1,...,g_{k-1})$, with \textit{ideal} defined as in abstract algebra (see Appendix~\ref{sec:suppl_math_proofs}).
Hence, the ideal generated by $(g_1,...,g_{k-1}, g_{k})$ is strictly larger than that generated by $(g_1,...,g_{k-1})$ alone, because we have added at least one new element (the $g_k$). We prove that in our setting the process of adding new independent $g_k$s will terminate (proof in Appendix~\ref{sec:suppl_math_proofs}):
\begin{theorem}
\label{thm:noetherian} 
When using Definition~\ref{def:restricted_indep} for independence and real-analytic functions to approximate $g$s, the process of  starting with a relation $g_1$ and iteratively adding new independent $g_k$s will terminate.
\end{theorem}

If $\M$ is real-analytic (i.e. is cut out by a finite set of equations of type $h(\tau)\!=\!0$ for some finite set of real-analytic $h$s), then after the process terminates, the set where all relations $g_1,..,g_k$ hold will be precisely $\M$.
Otherwise, the process will still terminate, having learned all possible analytic relations that hold on $\M$.
By a theorem of Akbulut and King \cite{akbulut1992approximating} any smooth submanifold of $\mathbb{R}^N$ can be approximated arbitrarily well by an analytic set, so in practice the differences would be negligible.

To ensure that each new relation decreases the data manifold dimension, we could additionally prohibit $g_1,..., g_k$ from having any syzygy $\{f_1,...,f_k\}$ in which $f_k$ itself is not expressible in terms of $g_1,..., g_{k-1}$.
With such definition (below) we could guarantee that a sequence of independent relations $g_1,...,g_k$ restricts the data to a submanifold of codimension at least $k$ (Theorem~\ref{thm:codim}, which we prove in Appendix~\ref{sec:suppl_math_proofs}).

\begin{definition}[Strong Independence]
\label{def:strong_indep}
$g_k$ is  strongly independent from $g_1,...,g_{k-1}$  if the equality 
$\sum_{j=1}^k f_j g_j \!=\! 0$
implies that $f_k$ is expressible as $f_k=h_1 \cdot g_1 + ... + h_{k-1} \cdot g_{k-1}$.
\end{definition}

\begin{theorem}
\label{thm:codim}
Suppose $g_1, \ldots, g_k$ is a  sequence of analytic functions on $B$, each strongly independent of the previous ones. Denote by  $\M_{\mathring{B}}=\{x\in \mathring{B}| g_j(x)=0 \text{ for all } j\}$ the part of the learned data manifold lying in the interior of $B$. Then dimension of $\M_{\mathring{B}}$ is at most $N-k$.
\end{theorem} 

In addition, we construct an alternative approach with similar dimensionality reduction guarantees, which ensures that the learned relations differ to first order. For this we use a notion of independence based on \textit{transversality}, with the following definition and lemmas (with proofs in Appendix~\ref{sec:suppl_math_proofs}):

\begin{lemma}
Dependence as in Definition~\ref{def:restricted_indep} implies  $\nabla\!_{\tau} g_k$ and $\nabla\!_{\tau} g_1,..., \nabla\!_{\tau} g_{k-1}$ are dependent.
\end{lemma}

\begin{definition}[Transversality]
\label{def:transverse_indep}
If for all points $\tau^{(i)} \!\in\! \M$ the gradients of $g_1,..,g_k$ at $\tau$, i.e. $\nabla_{\tau} g |_{\tau^{(i)}}$, are linearly independent, we say that $g_k$ is transverse to the previous relations: $g_k \pitchfork g_1,...,g_{k\text{-}1}$.
\end{definition}
\vspace{-5px}

Using transversality, we deem $g_k$ to be independent from $g_1,...,g_{k-1}$ if  the gradients of $g_k$ do not lie in the span of gradients of $g_1,...,g_{k-1}$ anywhere on $\M$. With this, $g_k$ that only differs from previous relations in higher-order terms would be deemed as `not new'.
This formulation is natural from the perspective of differential geometry. Let $H_{g_j}$ be the hypersurface defined by $g_j$: the set of points where $g_j\!=\!0$. Each $H_{g_1},...,H_{g_k}$ contains $\M$. If gradients of $g_k$ are linearly independent from gradients of $g_1,...,g_{k-1}$, then the corresponding hypersurfaces intersect transversely along $\M$.

\begin{lemma}
For once differentiable $(g_1,..,g_k)$ s.t. $H_{g_j}$s are transverse along their common intersection $H$, this intersection $H$ is a submanifold of $\R^N$ of dimension $N\!-\!k$.
\end{lemma}
\vspace{-5px}

The notion of independence defined via transversality is infinitesimal and symmetric w.r.t. permuting $g_k$s. This is useful in settings where many relations could be discovered, because it is then better to find relations whose first order behavior differs.
In cases where guaranteed decrease in dimension is not needed, using restricted syzygies could allow a flexible search for more expressive relations.

\subsection{Learning Latent Relations}
\label{sec:aml_algo}

\begin{figure}
\centering
\begin{subfigure}{0.48\textwidth}
\begin{algorithm}[H]
  \DontPrintSemicolon
  \SetAlgoCaptionSeparator{}
  \SetCustomAlgoRuledWidth{0.48\textwidth}
  $\{\tau^{(i)}\}_{i=1}^{d} \lar$ rollouts from RL actors\;
  train $g_1$ with loss from Eq.\ref{eq:g_loss}\;
  \For{$k = 2, 3, ..., $}{
    \If{aiming\_for\_transversality}{
    	train $g_k$ with loss $L_{tr}$ from Eq.\ref{eq:transverse_loss} \;
    }
    \Else(\tcp*[h]{using syzygies}){
    train $g_k$ with loss $L$ from Eq.\ref{eq:g_loss}\;
    \For{$j = 1, 2, ..., $}{
        generate $\tau_{o\!f\!f}, \tau_{o\!f\!f}^{test}$\;
        train $\mf_j$ with $L_{\mf} = |\mf_j(\tau_{o\!f\!f})|$\;
        \If{$\mf_j \!\neq\! 0$ on $\tau_{o\!f\!f}^{test}$}{
            break \tcp*[h]{$\!\!g_k \!\!\approx\!\!$ indep.}}
        \While{$\mf_j(\tau_{o\!f\!f}^{test}) \approx 0$}{
           freeze $\mf_j$\;
           train $g_k$ with $L_{syz}$ from Eq.\ref{eq:syz_loss}
       }
       \vspace{-3px}    
    }  
    \vspace{-3px}    
    } 
  \vspace{-3px}
  } 
  \caption{\hspace{-2px}\textbf{Algorithm 1:}Analytic Manifold Learning \small{\textsc{(aml)}}}
\vspace{-3px}
\end{algorithm}
\end{subfigure}
\hspace{5px}
\begin{subfigure}{0.48\textwidth}
\includegraphics[width=0.98\textwidth]{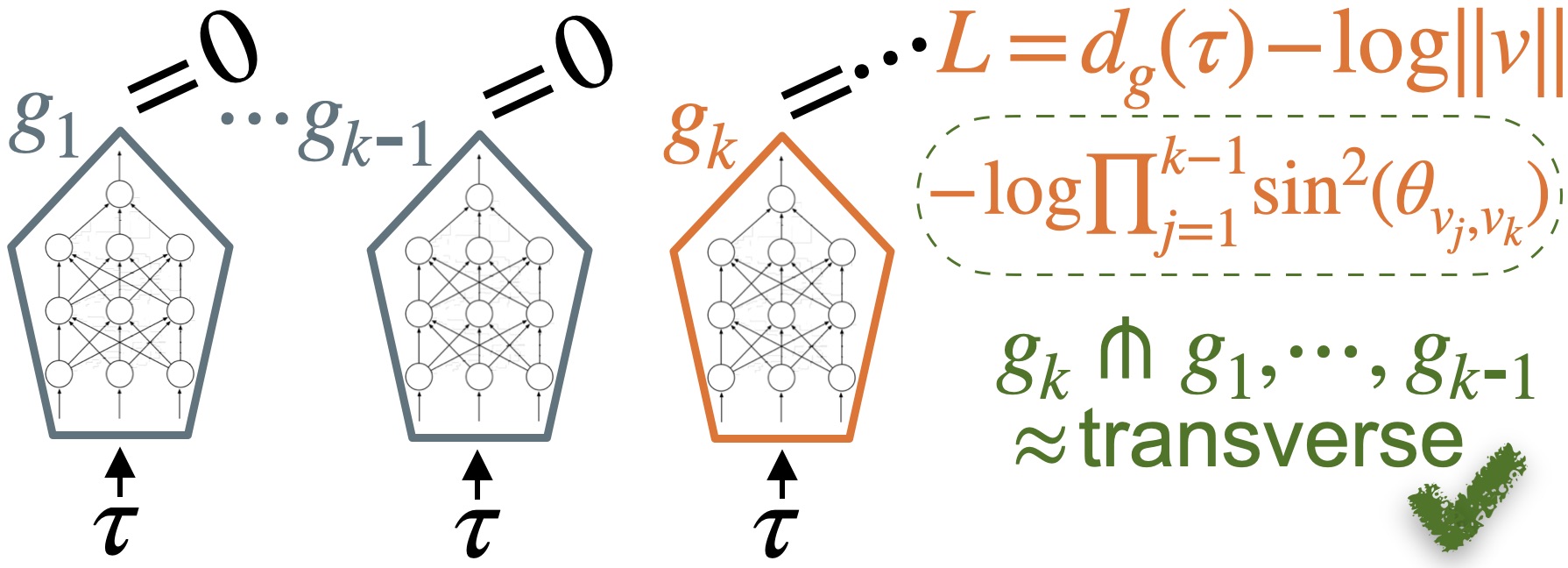}
\includegraphics[width=0.98\textwidth]{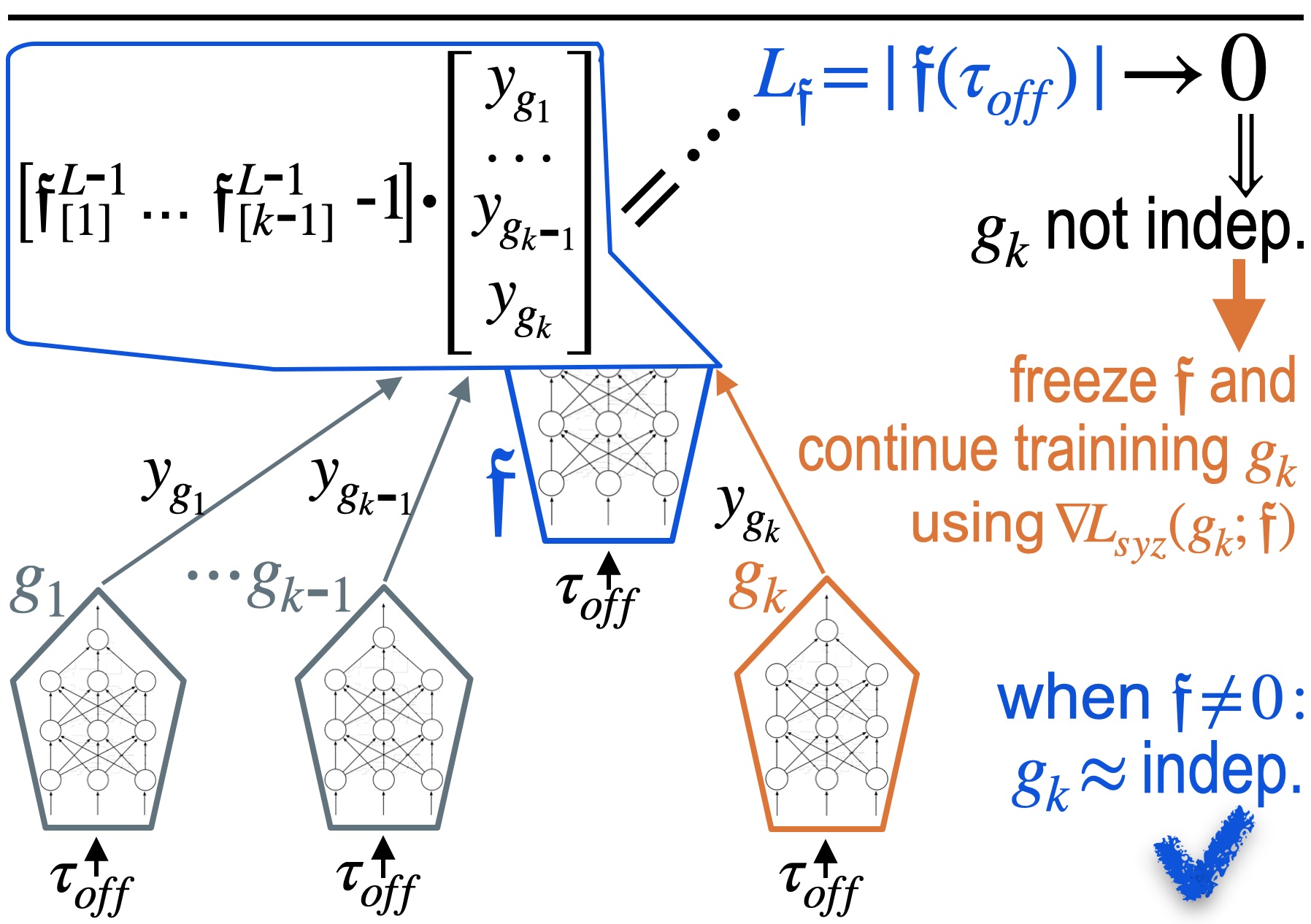}
\end{subfigure}
\vspace{-4px}
\caption{Left: algorithm for learning latent relations. Top right: using transversality. Bottom right: training with syzygy $\mf$ to uncover if $g_k$ is dependent, then using $\mf$ to modify $g_k$'s loss. Orange \& blue denotes NNs whose weights are being trained. Gray denotes learned relations whose NNs are frozen.}
\label{fig:algo}
\vspace{-5px}
\end{figure}

Here we describe the algorithm with relations $g_k$ and restricted syzygies $\mf$ approximated by neural networks. 
Each $g$ is represented by a neural network (NN) that takes a sequence of latent/low-dimensional states $\tau \!=\! [s_t, s_{t+1},...,s_T], \tau \!\in\!\R^N $ as input. The output of $g$ is a scalar. We use $g$ to denote both the relation and the NN used to learn it. If $g$ outputs 0 for on-manifold data, this implies $g$ has learned a function $g(\tau)\!=\!0$, which captures a relation between states of the underlying dynamical system. 
$g$ is trained on minibatches of size $b$ of on-manifold data points $\tau^{(i)}$ using loss gradients: $\nabla\!L \!=\! \textstyle\sum_{i=1}^{b} \!\nabla\!_{g} \big[ L(\tau^{(i)})\big]$, where $\nabla\!_{g}$ means gradient w.r.t NN weights of $g$. We need to make $g \!\rar\! 0$ for on-manifold data, while avoiding trivial relations (e.g. all NN weights $\approx\! 0$). Hence, in the loss we minimize $d_g(\tau)\!=\!\frac{|g(\tau)|}{\norm{v}}$, where $v$ is the gradient of $g$ with respect to input points $\tau^{(i)}$: $v \!=\! \nabla\!_{\tau}(g)|_{\tau^{(i)}}, v \!\in\!\R^N$. The gradient norm $\norm{v}$ is the maximal `slope' of the linearization of $g$ at $\tau$, so $d_g(\tau)$ is the distance from $\tau$ to the nearest point where this linearization vanishes ($d_g(\tau)\!=$~height/slope~$=$~distance). Hence, $d_g(\tau)$ is a proxy for the distance from $\tau$ to the vanishing locus of $g$. 
This measure of vanishing avoids scaling problems (see Appendix~\ref{sec:suppl_math_related_work}).
We also maximize $\log\norm{v}$ to further regularize $g$. Equation~\ref{eq:g_loss} summarizes our loss for $g$:
\begin{align}
L(g) = d_g(\tau) - \log \norm{v} \ ; 
\quad d_g(\tau) = |g(\tau)|/\norm{v} \ ;
\quad v = \nabla\!_{\tau}(g)|_{\tau}
\label{eq:g_loss}
\end{align}
We proceed sequentially: first learn $g_1$, then $g_2$, and so on.
Suppose that so far we learned (approximately) independent relations $g_1,...,g_{k-1}$. We then keep their NN weights fixed and learn an initial version of the next relation $g_k$.
To obtain $g_k$ that is transverse to $g_1,..,g_{k-1}$ (Definition~\ref{def:transverse_indep}), we augment the loss as follows.
We compute gradients of each $g_1,...g_{k-1}$ w.r.t input $\tau$. For example, for $g_1$ we denote this as $v_1 \!=\! \nabla\!_{\tau}(g_1)|_{\tau}$. 
Making $g_k$ transverse to $g_1,...g_{k-1}$ means ensuring that $v_k$ is linearly independent of $v_1,...,v_{k-1}$.
We optimize a computationally efficient numerical measure of this:
maximize the angles between $v_k$ and all the previous $v_1,..,v_{k-1}$. Such measure encourages transversality of subsets of relations and strongly discourages small angles. Our overall measure of transversality is the product of sines of pairwise angles, with log for stability (Appendix~\ref{sec:suppl_b31} gives further discussion):
\vspace{-5px}
\begin{align}
L_{tr}(g_k) = d_{g_k}(\tau) - \log\norm{v_k} - \log \textstyle\prod_{j=1}^{k-1} \sin^2(\theta_{v_j,v_k})
\label{eq:transverse_loss}
\end{align}
For independence based on Definition~\ref{def:restricted_indep}, we instead learn a restricted syzygy $\mf(\tau_{o\!f\!f}, g_1,...,g_k) = 0$.
Training data for $\mf$ is comprised of: 1) $\tau_{o\!f\!f}$ (defined in Section~\ref{sec:math})
and 2) $y_{g_1}\!\!=\!\!g_1(\tau_{o\!f\!f}), ...,  y_{g_k}\!\!=\!\!g_k(\tau_{o\!f\!f})$, i.e. outputs from $g_1,...g_k$ with $\tau_{o\!f\!f}$ fed as inputs. $y_gs$ are passed directly to the next-to-last layer, which we denote as $\mf^{L\text{-}1} \!\in \R^{k-1}$.
The last layer of $\mf$ computes a dot product of $\big[\mf^{L\text{-}1}_{[1]},...,\mf^{L\text{-}1}_{[k\text{-}1]}, \text{-}1\big]$ and $[y_{g_1},...,y_{g_k}]$. We use a simple L1 loss for training $\mf$.
If $\mf$ outputs 0 at convergence: $g_k$ is not independent.
In this case, we freeze the weights of $\mf$ and continue to train $g_k$ with augmented loss. We use gradients passed through $\mf$ to push $g_k$ away from a solution that made it possible to learn $\mf$:
\vspace{-10px}
\begin{align}
\label{eq:syz_loss}
\!\nabla\!L_{syz}(g_k ; \mf) = \nabla\!L(g_k) - \!\nabla\!_{g_k} \Big[ \big|\mf(\tau_{o\!f\!f},g_1,...,g_k)\big| \Big]
\end{align}
\vspace{-1px}
$L_{syz}$ encourages adjusting $g_k$ such that it makes the outputs of (frozen) $\mf$ non-zero. Once $L_{syz}(g_k; \mf)$ is minimized, we can attempt to learn another syzygy $\mf_2$, and so on, until we cannot uncover any new dependencies. 
Then $g_k$ can be declared (approximately) independent of $g_1,...g_{k-1}$ and we can proceed to learn $g_{k+1}$. All $g_k$s, $\mf$s, $L$s are in latent space, so networks are small \& quick to train.

An additional benefit of our formulation is that prior knowledge can be incorporated without restricting the hypothesis space. $g_k$s can be pre-trained in a supervised way: to output values that a prior heuristic produces on- and off-manifold. Then, $g_k$s can be further trained using on-manifold data, and if prior knowledge is wrong, then $g_k$ would move away from the wrong heuristic during further training.

\section{Imposing AML Relations During Transfer}
\label{sec:aml_transfer}

The previous section described how to encode the latent data manifold onto a set of analytic relations represented by neural networks. This section shows how to impose these relations into a latent space of a sequential VAE.

\begin{wrapfigure}{r}{0.25\textwidth}
\vspace{-25px}
\includegraphics[width=0.25\textwidth]{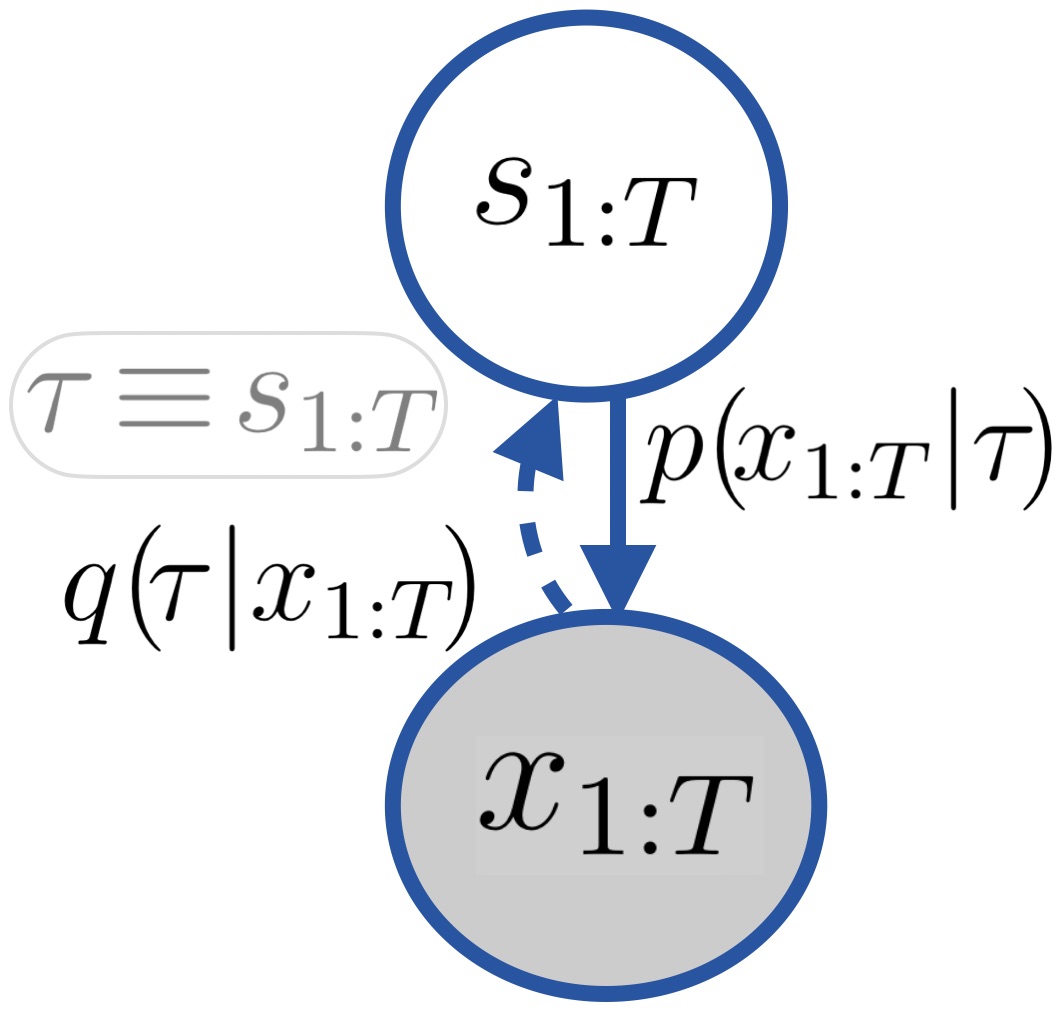}
\vspace{-17px}
\caption{\small{\mbox{A basic sVAE}}}
\label{fig:thesis_svae_very_simple}
\vspace{-20px}
\end{wrapfigure}
The model for a basic sequential VAE could be defined as follows:
\vspace{-3px}

$\text{Generative model: } p(\tau, x_{1:T}) := p(s_{1:T})  \textstyle p(x_t | x_{<t}, s_{1:T})$
\vspace{-3px}

$\text{Approximate posterior: } q(\tau | x_{1:T}) := q(s_{1:T} | x_{1:T})$
\vspace{-3px}

To model long-term dependencies it is customary to use recurrent neural networks (e.g. with LSTM, GRU, or other recurrent units). However, in case of non-stationary data these can slow down or even stagnate VAE's learning progress. Hence, a better choice is to use a convolutional encoder and decoder, with a one or more fully connected layers at the bottleneck. 

Several recent works showed that a further improvement can be achieved by requiring to predict $L$ next frames instead of only reconstructing the given frames~\cite{yin2017hashing, zintgraf2019varibad}. Hence, we use a simple predictive version of a sequential VAE that we call $P\!RED$. Section~\ref{sec:bench} benchmarked $P\!RED$ against other VAE variants and demonstrated that, despite its simple architecture, $P\!RED$ outperformed sequential VAE approaches with more sophisticated recurrent architectures.

\begin{itemize}[leftmargin=0.4cm]
\item[]$\pmb{P\!RED}$: a VAE that, given a sequence of frames $x_{1},...,x_{T}$, constructs a predictive sequence $x_{1},...,x_{T\text{+}L}$. First, the convolutional stack is applied to each $x_t$; then, the $T$ output parts are aggregated and passed through fully connected layers. Their output constitutes the predictive latent state. To decode: this state is chunked into $T\text{+}L$ parts, each fed into deconvolutional stack for reconstruction.
\end{itemize}
\vspace{1px}

\noindent AML relations can be imposed on the latent state of $P\!RED$ by augmenting the latent part of the loss as follows:
\begin{align}
\begin{split}
\mathcal{L}_{P\!RED}^{AML} \!=\! \E\!\!\!_{\substack{\tilde{\btau}_{1:\tpl} \sim \\ q(\tau_{1:\tpl}|x_{1:T})}}\Big[ &-\!\Big(\overbrace{\log p({\color{darkmagenta}x_{1:\tpl}}|\tilde{\tau}_{1:\tpl}\!) \!-\! K\!L\big(q||\mathcal{N}\!(0,\!1\!)\big)}^{\text{standard } ELBO \text{ for } P\!RED \text{ version of } V\!AE}\Big)
\\
&+\underbrace{\textstyle\sum_{k=1}^K \big|g_k(\tilde{\tau}_{1:\tpl},a_{1:\tpl})\big|}_{\text{impose } {\ouralgo} \text{ relations}} \ \Big]
\label{eq:aml_elbo}
\end{split}
\end{align}
In the above, $\tilde{\tau}_{1:\tpl}$ denotes a sample from the approximate posterior $q(\tau_{1:\tpl}|x_{1:T})$. $p({\color{darkmagenta}x_{1:\tpl}}|\tilde{\tau}_{1:\tpl}\!)$ denotes the likelihood for $P\!RED$, with magenta color indicating that decoder outputs a predictive sequence ${\color{darkmagenta}\hat{x}_{1:\tpl}}$ instead of a reconstruction $\hat{x}_{1:t}$. 
We would like to validate that imposing AML relations works well when the data is non-stationary, hence we train $P\!RED$ on data stream generated during RL training. In this case, instead of learning relations on a subsequence of states, during AML training (described in the previous section) we learn relations on subsequences that include actions: $\tau = [s_1,a_1,s_2,a_2,...]$.
\\
\\

\vspace{-25px}
\section{Evaluating {\ouralgo} and Latent Space Transfer}
\label{sec:aml_experiments}

\begin{figure}[t]
\includegraphics[width=0.15\textwidth]{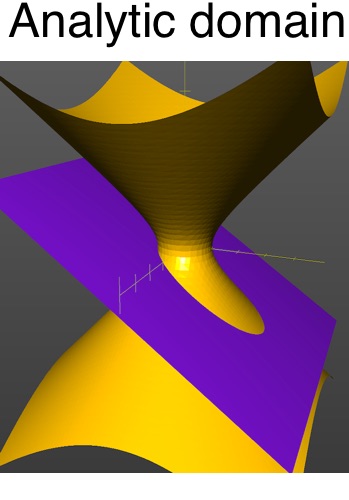}
\hspace{5px}
\includegraphics[width=0.18\textwidth]{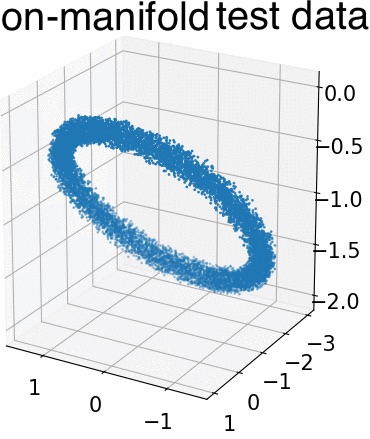}
\hspace{10px}
{\color{gray}\vrule}
\hspace{10px}
\includegraphics[width=0.50\textwidth]{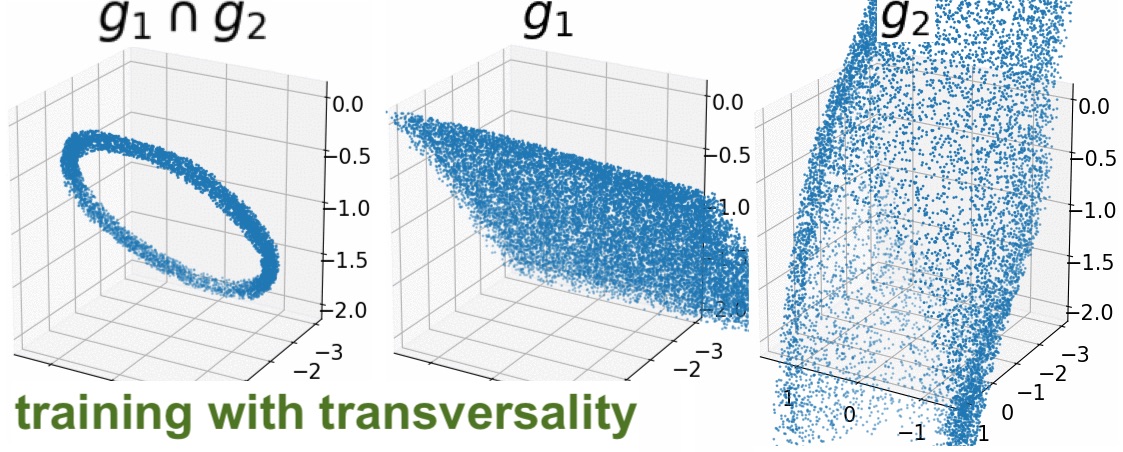}
\vspace{1px}
{\color{gray}\hrule}
\vspace{1px}
\includegraphics[width=0.82\textwidth]{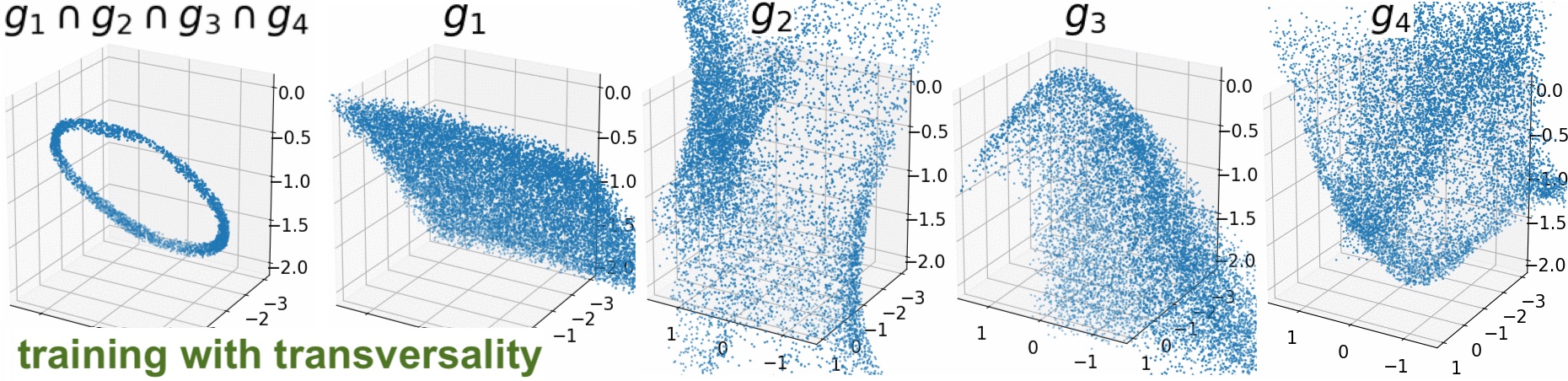}
\vspace{1px}
{\color{gray}\hrule}
\vspace{1px}
\includegraphics[width=0.99\textwidth]{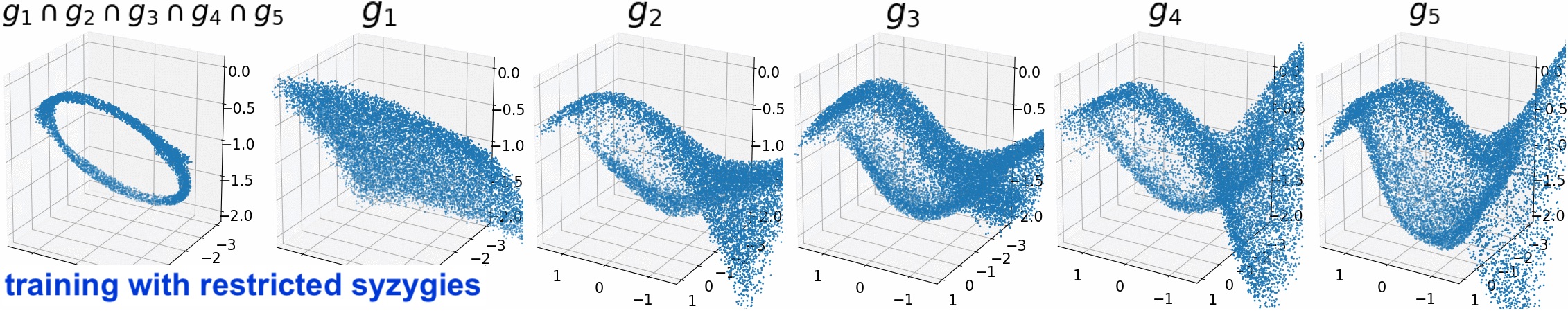}
\caption{Top row: visualization of the analytic domain and noisy on-manifold data, followed by AML relations trained using transversality. First, $g_1 \cap g_2$ is shown: this is the intersection of the learned relations (i.e the intersection of the zero-level sets of $g_1,g_2$). The \mbox{zero-level} sets of individual relations are shown next. With transversality we get $g_1 \cap g_2$ as two simple relations: a plane and a hollow cylinder.
Middle row: AML using transversality when we insist on getting more than 2 relations: $g_1 \cap g_2 \cap g_3 \cap g_4$ then includes smoothed cones.
Bottom row: AML using restricted syzygies yields more advanced shapes.}
\label{fig:aml_analytic_results}
\vspace{-25px}
\end{figure}

We evaluate the proposed {\ouralgo} approach with 3 sets of experiments: 1) learning on an analytic domain and visualizing relations in 3D; 2) handling dynamics with friction and drag on a \textit{block-on-incline} domain; 3) employing relations learned on a source domain to get better latent space properties on the \textit{YCB-on-incline} as target.

\vspace{-7px}
\subsection{Evaluating AML Training}

Figure~\ref{fig:aml_analytic_results} visualizes the results on the analytic domain, for which on-manifold data comes from an intersection of a hyperboloid and a plane. It illustrates that using AML with transversality allows us to capture the latent data manifold using a small number of general relations, i.e. relations with simple shapes. The illustrations in 3D make it easy to see that these intersect transversely (top row), or attempt to maximize the angle of intersection (middle row). In contrast, relations found using syzygies have more complicated shapes and can be similar in some regions, as expected (bottom row). This could be useful when we need to avoid large changes, e.g. for fine-tuning or for flexible partial transfer using subsets of relations.

\begin{wrapfigure}{r}{0.30\textwidth}
\centering
\vspace{-22px}
\includegraphics[width=0.30\textwidth]{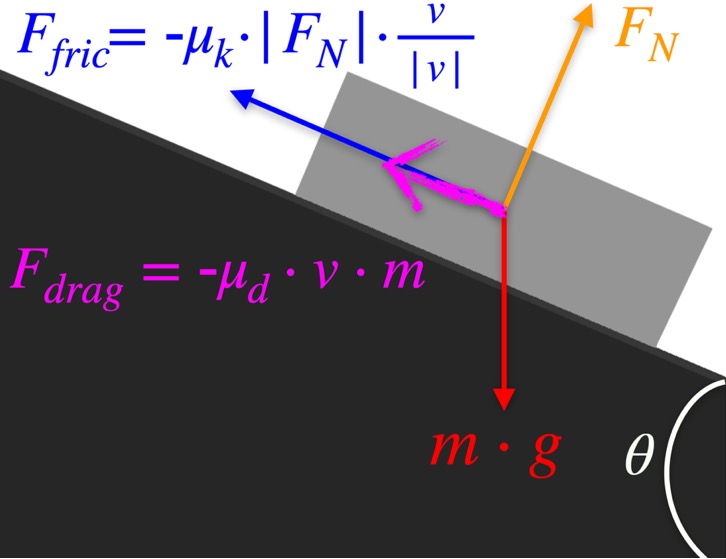}
\vspace{-15px}
\caption{\textit{block-on-incline}}
\label{fig:block_on_incline}
\vspace{-8px}
\end{wrapfigure}
\begin{figure}[t]
\hspace{1px}
\includegraphics[width=0.213\textwidth]{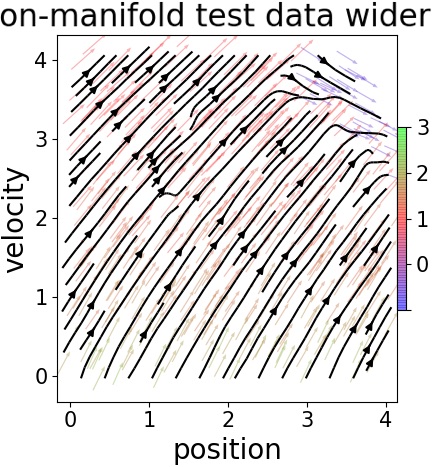}
\hspace{4px}
\includegraphics[width=0.17\textwidth]{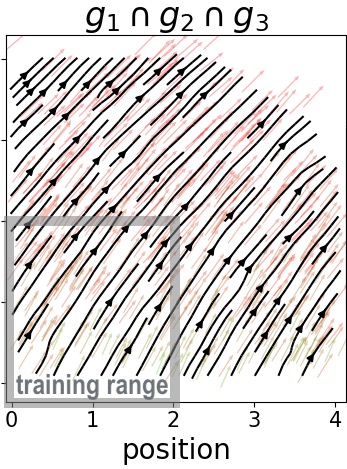}
\hspace{6px}
\includegraphics[width=0.17\textwidth]{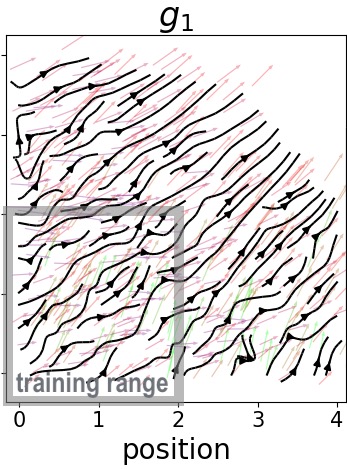}
\includegraphics[width=0.17\textwidth]{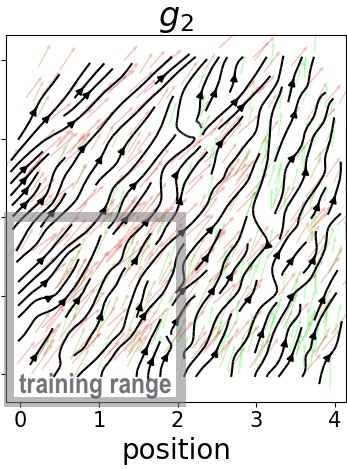}
\includegraphics[width=0.17\textwidth]{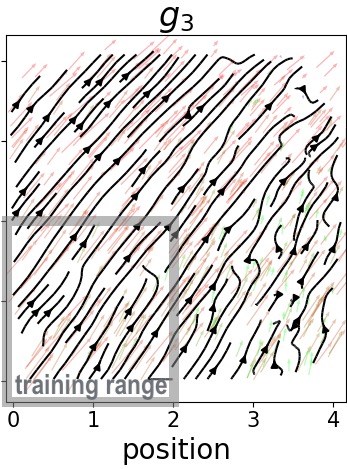}
\\
\includegraphics[width=0.225\textwidth]{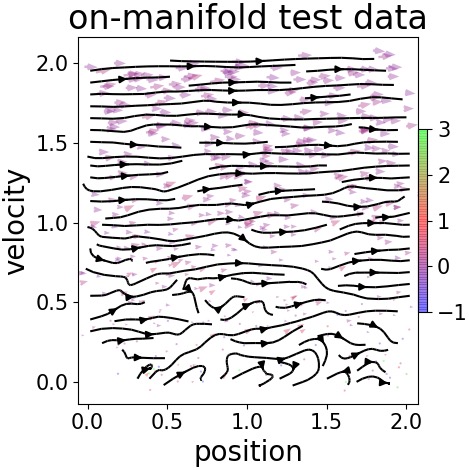}
\hspace{3px}
\includegraphics[width=0.17\textwidth]{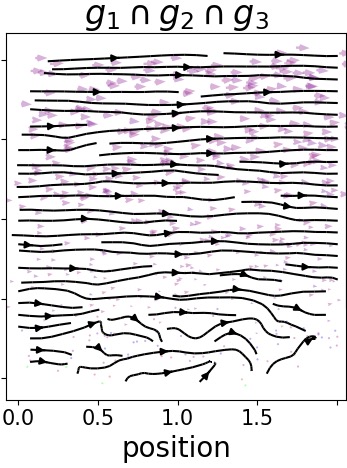}
\hspace{5px}
\includegraphics[width=0.17\textwidth]{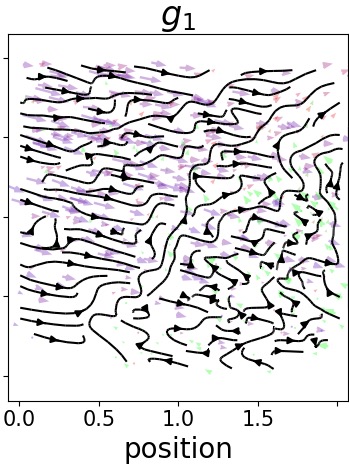}
\includegraphics[width=0.17\textwidth]{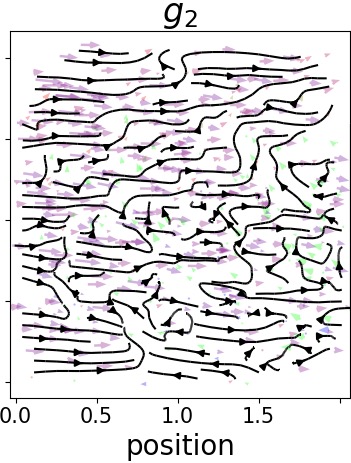}
\includegraphics[width=0.17\textwidth]{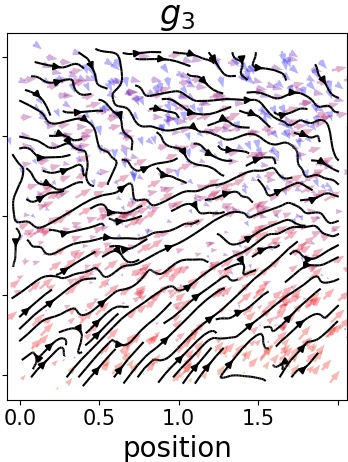}
\\
\includegraphics[width=0.225\textwidth]{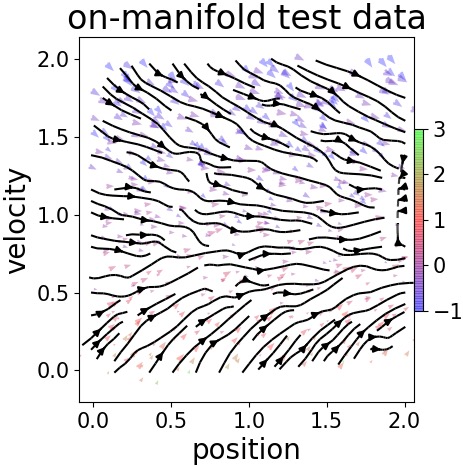}
\hspace{3px}
\includegraphics[width=0.17\textwidth]{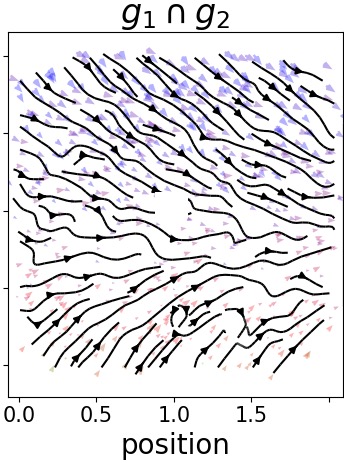}
\hspace{5px}
\includegraphics[width=0.17\textwidth]{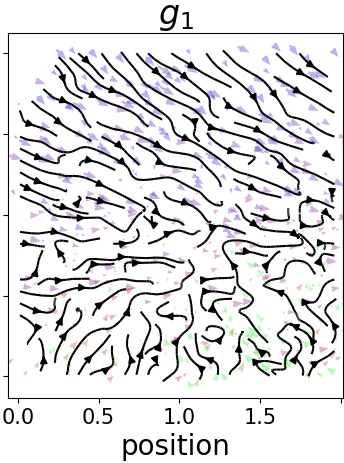}
\includegraphics[width=0.17\textwidth]{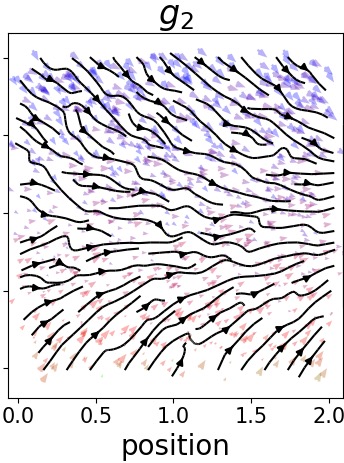}
\caption{Phase space plots for \textit{block-on-incline} domain. 1st column: on-manifold data; 2nd column: the learned manifold encoded by the intersection of AML relations trained with transversality; rest: zero-level sets of individual AML relations. Arrows show change in position \& velocity after $1sec$ of sliding (scaled to fit). Top row: plots show the case of a $45^{\circ}$ incline and demonstrate generalization. {\ouralgo} is only given training data with start position \& velocity $\in [0,0.2]$, but is able to generalize to $[0,0.4]$.
Middle row: high friction on a $35^{\circ}$ incline. Bottom row: high drag on a $10^{\circ}$ incline. }
\label{fig:aml_block_transverse}
\vspace{-20px}
\end{figure}

Next, we evaluate {\ouralgo} on a physics domain: a block sliding down an incline (Figure~\ref{fig:block_on_incline}).
The block is given a random initial velocity; gravity, friction and drag forces then determine its further motion. On-manifold data consists of noisy position \& velocity of the block at the start and end of trajectories.
Figure~\ref{fig:aml_block_transverse} shows results for {\ouralgo} with transversality.
Appendix~\ref{sec:suppl_b33} gives results with syzygies.
Overall, these results show that {\ouralgo} can generalize beyond training data ranges and capture non-linear dynamics. \hfill \linebreak
\\

\vspace{-2px}
\subsection{Latent Space Transfer}

\begin{wrapfigure}{r}{0.45\textwidth}
\centering
\vspace{-10px}
\includegraphics[width=0.45\textwidth]{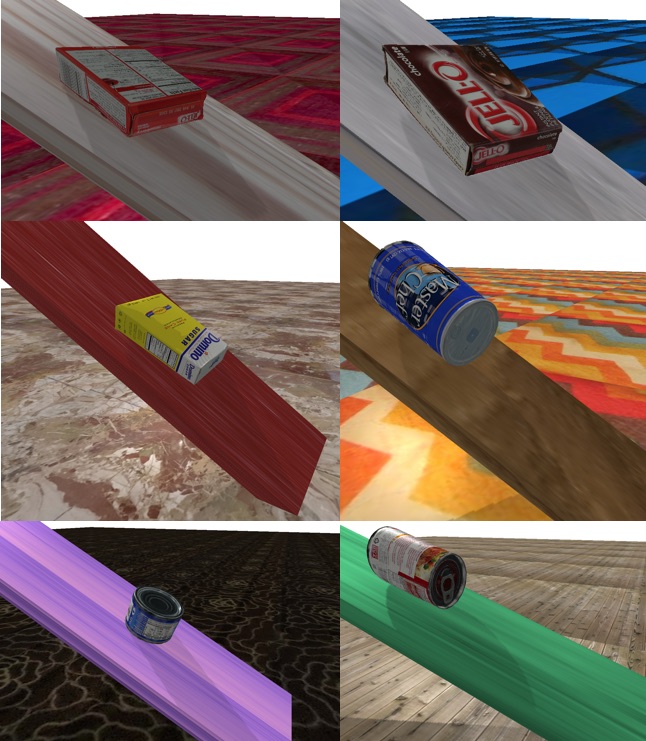}
\caption{\textit{YCB-on-incline} domain.}
\label{fig:aml_ycb_on_incline_domain}
\vspace{-10px}
\end{wrapfigure}
\textbf{Emulating sim-to-real with sim-to-sim:} In the final experiment we use a sim-to-sim setting that attempts to emulate challenges that arise in sim-to-real problems. We use a simulator with simple geometric shapes that move along an incline as a source domain. We refer to this source/simulation domain as \textit{Geom-on-incline}.
Since in sim-to-real settings we have access to the simulation state (robot and objects' positions, velocities, etc): we will use this simulation state of \textit{Geom-on-incline} directly (instead of using RGB images from the source domain).
To emulate a `real' domain, we use an advanced simulation involving objects with real mesh scans and RGB images as observations. We refer to this domain as \textit{YCB-on-incline} (visualized in Figure~\ref{fig:aml_ycb_on_incline_domain}), since it uses objects from the YCB dataset~\cite{calli2015ycb}. \textit{YCB-on-incline} yields realistic visual appearances and non-trivial object dynamics. The dynamics is dictated by meshes obtained from the 3D scans of real objects. Hence, there is a non-trivial sim-to-`real' gap between the dynamics of the simple shapes of \textit{Geom-on-incline} domain vs realistic shapes of the \textit{YCB-on-incline} domain.
Furthermore, instead of building a dataset of RGB images, we train VAE-based learners directly on the stream of RGB frames that an RL agent generates during its own training. Hence, we ensure that the distribution of observations obtained on the target domain is non-stationary.
\vspace{5px}

\noindent \textbf{Measuring encoder distortion:}
One important quality measure of a latent space mapping is how much it distorts the true data manifold. For the experiment that runs on a sim-to-sim setting we have access to the low-dimensional states of the simulator of the target domain. We denote such sequences of low-dimensional simulation states as $\tau^{true}$. We then quantify the distortion of the encoder map (on 10K test points) as follows: we take pairs of low-dimensional representations $\tau_1^{true}$, $\tau_2^{true}$ and the corresponding pixel-based representations $x_1, x_2$, then compute distortion coefficient $\rho_{distort}$:
\begingroup
\setlength{\abovedisplayskip}{3pt}
\setlength{\belowdisplayskip}{5pt}
\setlength{\abovedisplayshortskip}{3pt}
\setlength{\belowdisplayshortskip}{5pt}
\begin{align}
\rho_{distort} &= \log d_{L2}\big(\phi_{enc}(x_1), \phi_{enc}(x_2)\big) \big/ d_{L2}\big(\tau_1^{true}, \tau_2^{true}\big)
\label{eq:aml_distortion_measure}
\end{align}
\endgroup
Here, $d_{L2}$ is the Euclidean distance. An encoder that yields low \textit{variance} of these coefficients preserves the geometry of the low-dimensional manifold better (up to overall scale). This measure is related to approaches surveyed in~\cite{distort18, bartal2019dimensionality} (see Appendix~\ref{sec:suppl_b32}).
The above evaluation could be done on an actual sim-to-real setting as well, if the real part is equipped with additional sensing. For example: if robot's joint angles and velocities are accurately reported; object positions and orientation are accurately estimated by a motion capture system. Such evaluation would be feasible in well-equipped labs. However, a sim-to-sim evaluation could be enough to compare the quality of the encoders produced by various algorithms, so doing this evaluation in an actual sim-to-real setting is not strictly required.
\vspace{5px}

\noindent \textbf{Latent Space Transfer with AML:}
In this experiment, we compare the results of imposing AML relations when training $P\!RED$ on \textit{YCB-on-incline} to two baselines: 1) $P\!RED$ without AML relations imposed, and 2) a basic non-sequential $V\!\!AE$.

First, {\ouralgo} learns relations from \textit{Geom-on-incline} domain. Incline angle, friction and object pose are initialized randomly.
Actions are random forces that push objects along the incline. {\ouralgo} is given the incline angle, position \& velocity at two subsequent steps, and the applied action. Note that, for the sim-to-real and sim-to-sim settings we have access to the simulator state of the source domains. Hence, we can use the low-dimensional state sequences to as training data for AML to learn the `latent' data manifold of the source domain. For general cases that do not involve physics simulators: we could instead learn a VAE-based embedding on the source domain. This would yield the low-dimensional latents for learning AML relations on the source domain.

\begin{figure}[b]
\vspace{-10px}
\centering
\includegraphics[width=0.40\textwidth]{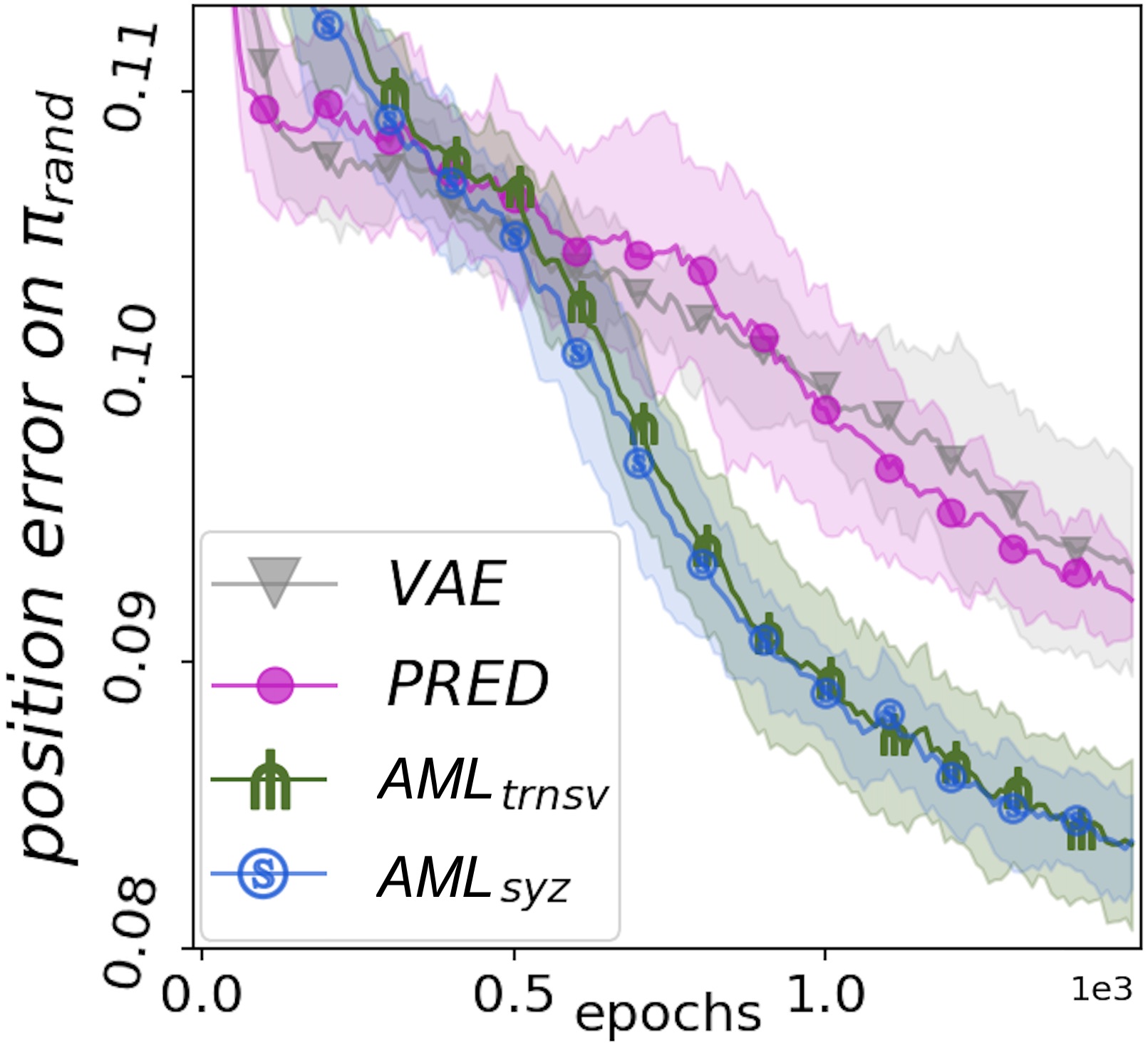}
\includegraphics[width=0.40\textwidth]{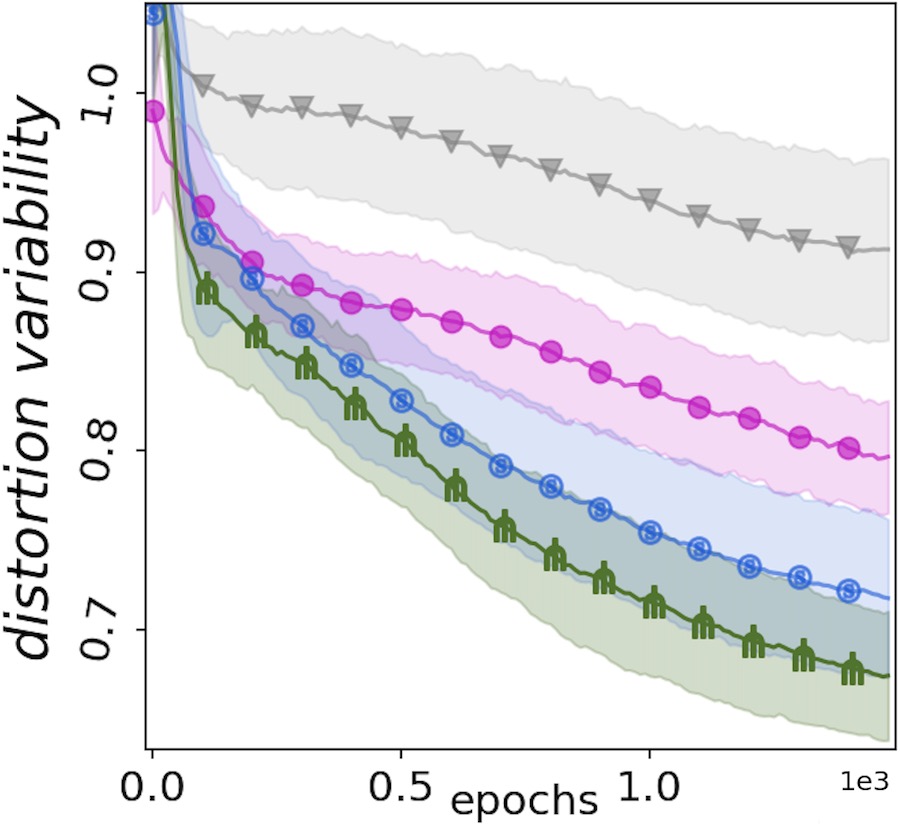}
\caption{Results of imposing AML relations trained from \textit{Geom-on-incline} domain when training $P\!RED$ on the target \textit{YCB-on-incline}. Left plot shows latent state alignment for object position computed using evaluation suite from Section~\ref{sec:suite}. The test observations for evaluation are obtained using a random policy $\pi_{rand}$. This is because we want to ensure that the alignment is good everywhere, not just in the state space regions that are visited often by the current RL policy. The results indicate that {\ouralgo}$_{\text{trnsv}}$ and {\ouralgo}$_{\text{syz}}$ yield better alignment.
The right plot in Figure~\ref{fig:aml_sliding_ycb_results}
shows the distortion variability of the encoder map during training. It indicates that
{\ouralgo}$_{\text{trnsv}}$ and {\ouralgo}$_{\text{syz}}$ produce less distorting encoders that those obtained by than $V\!\!AE$ and $P\!RED$ without {\ouralgo} relations imposed.}
\label{fig:aml_sliding_ycb_results}
\end{figure}

Then, we evaluate the latent space transfer to the target domain. For this, we train an unsupervised learner ($P\!RED$) on the target domain: \textit{YCB-on-incline}. This domain is visualized in Figure~\ref{fig:aml_ycb_on_incline_domain}.
Recall that this sim-to-sim setup aims to emulate the challenges of sim-to-real transfer. In this case, \textit{Geom-on-incline} plays a role of a simulator, while RGB frames from \textit{YCB-on-incline} act as surrogates for `real' observations. The distribution of the frames is non-stationary, since they are sampled using the current (changing) policy of the RL learner.
We use PPO~\cite{schulman2017proximal} to learn RL policies.
The reward for RL is defined to be proportional to how close an object stays to the middle of the incline, i.e. RL has to learn to counteract gravity without pushing objects off the narrow incline plane. This task is challenging, since several YCB objects we use have rounded shapes, causing them to easily roll off the sides of the incline plane. The textures of the ground and the incline plane are initialized randomly at the start of each episode.

We impose {\ouralgo} relations by extending the latent part of an ELBO-based loss as defined by Equation~\ref{eq:aml_elbo}, explained in detail in the previous section.
The left plot in Figure~\ref{fig:aml_sliding_ycb_results}  shows that the resulting {\ouralgo}$_{\text{trnsv}}$ ({\ouralgo}$_{\text{syz}}$ when using syzygies) gets a better latent state alignment for object position, compared to $V\!\!AE$ and $P\!RED$ without {\ouralgo} relations imposed.
The right plot in Figure~\ref{fig:aml_sliding_ycb_results} confirms that imposing AML relations trained with either transversality ($AML_{trnsv}$) or syzygies ($AML_{syz}$) yields an encoder with a lower distortion variability (i.e. lower variance of $\rho_{distort}$ coefficients defined by Equation~\ref{eq:aml_distortion_measure}).

Overall, results in Figure~\ref{fig:aml_sliding_ycb_results} show that imposing {\ouralgo} relations helps improve the latent space mapping of $P\!RED$ when training on RGB frames that come from a non-stationary data stream of an RL learner.

\section{Related Work}
\label{seq:related_work}

Scalable simulation suites for continuous control~\cite{ roboschool, tassa2018deepmind, pybulletgym} bolstered progress in deep RL. However, advanced benchmarks for unsupervised learning from non-stationary data are lacking, since the community mainly focused on dataset-oriented evaluation. \cite{anand2019unsupervised} provides such a framework for ATARI games, but it is not aimed at continuous control.
\cite{raffin18srl} includes a limited set of robotics domains and 3~metrics for measuring representation quality: KNN-based, correlation, RL reward. We incorporate more standard benchmarks, introduce a variety of objects with realistic appearances (fully integrated into simulation) and measure alignment to latent state in a complimentary way (highly non-linear, but not RL-based). In future work, it would be best to create a combined suite to support both games- and robotics-oriented domains, and offer a comprehensive set of RL-based and RL-free evaluation.

Our formulation of learning latent relations is in the general setting of representation learning. 
This is a broad field, so in this work we focus on formalization of learning independent/modular relations that capture the true data manifold. We also provide a way to transfer relations learned on source domains to target domains.
Unlike meta-learning, we do not assume access to a task distribution and do not view target task reward as the main focus.
Our sequential approach to learning $g_1,...,g_k$ has conceptual parallels with a functional Frank-Wolfe algorithm~\cite{jaggi2013revisiting}, but without convex optimization.
Learning sequentially helps avoid instabilities, e.g. from training flexible NN mixtures with EM~\cite{IODINE19}.
There is prior work for learning algebraic (meaning polynomial) relations, but its criterion for relation simplicity is based on polynomial degree.
Such approaches are based on computational algebra and spectral methods from linear algebra.
This line of work was initiated by~\cite{livni2013vanishing, sauer2007approximate, heldt2009approximate}, with
extensions~\cite{fassino2010almost, fassino2013simple, kera2016noise, kera2019spurious, kera2019gradient}, applications~\cite{iraji2017principal, yan2018deep} and learning theory analysis~\cite{hazan2016non, globerson2017effective}.
Our formulation is more general, since we learn analytic relations and approximate them with neural networks. We summarize the main differences \& point out potential connections in Appendix~\ref{sec:suppl_math_related_work}.

\section*{Conclusion and Future Work}

We proposed a suite for evaluation of latent representations and showed that additional latent space structure can be beneficial, but could stifle learning in existing approaches. 
We then presented {\ouralgo}: a unified approach to learn latent relations and transfer them to target domains. We offered a rigorous mathematical formalization, algorithmic variants \& empirical validation for {\ouralgo}.

We showed applications of {\ouralgo} to physics \& robotics domains. However, in general
{\ouralgo} does not assume that source or target domains are from a certain field, such as robotics, or have particular properties, such as continuity in the adjacent latent states or existence of an easy-to-learn transition model. As as long some relation exists between the subsequences of latent states -- {\ouralgo} would attempt to learn it, and would succeed if a chosen function approximator is capable of representing it.
Moreover, {\ouralgo} relations can be learned on the latent space of any unsupervised learner trained on the source domain. In this case, {\ouralgo} would capture abstract relations that encode the regularities embedded in the latent representation learned on the source domain. Imposing these relations during transfer could help to preserve (i.e. carry over) these regularities. This alternative could be better than starting from scratch and better than fine-tuning. Starting from scratch is not data-efficient. Fine-tuning is prone to getting stuck in local optima, causing permanent degradation of performance, especially in case of a non-trivial mismatch between the source and target domains.

{\ouralgo} can build a modular representation of relations encoded in the latent/low-dimensional space. Hence, {\ouralgo} can enable a dynamic partial transfer and thus help recover from negative transfer in cases of large source-target mismatch. In our follow-up work, we intend to dynamically adjust the strength of imposing each latent relation on the target domain. For this, we would combine the learned relations $g_1,...,g_k$ using prioritization weights $w_1,...,w_k$. These weights would be optimized by propagating the gradients of the RL loss w.r.t. the latent state representation (that these weights would influence). 
Further extensions could include, for example, lifelong learning: we could gradually expand the set of learned relations and discard relations whose weights decay to zero as the lifelong learning proceeds. 
Another promising option would be to learn policy representations (rather than state representations). If {\ouralgo} could be used to learn policies that are in some sense independent, then we could provide a way to learn a \textit{portfolio} of policies that are complementary. Then, we could construct algorithms for learning diversified portfolios, such that a system capable of executing any policy in a portfolio could provide robustness to uncertainty and changes in the environment.

\ack{
This research was supported in part by the Knut and Alice Wallenberg Foundation.
This work was also supported by an ``Azure for Research'' computing grant.
We would like to thank Yingzhen Li, Kamil Ciosek, Cheng Zhang and Sebastian Tschiatschek for helpful discussions regarding unsupervised  \& reinforcement learning and variational inference.
}

\appendix
\counterwithin{figure}{section}
\part*{\large{Appendices}}
\vspace{-15px}
\startcontents[sections]
\printcontents[sections]{l}{1}{\setcounter{tocdepth}{2}}
\vfill

\renewcommand{\thesection}{\Alph{section}}

\section{Evaluation Suite for Unsupervised Learning for Continuous Control}
\label{sec:suppl_a}
\setcounter{figure}{0}
\vspace{-10px}
\subsection{Benchmarking Alignment : Algorithm Descriptions and Further Evaluation Details }
\vspace{-7px}

In this section, we include more detailed descriptions of the existing approaches we evaluated, describe parameters used for evaluation experiments, and give examples of reconstructions.
Code for the evaluation suite environments can be obtained at:
{\footnotesize\url{https://github.com/contactrika/bulb}
}%

\vspace{1px}
\ \ -- $V\!\!AE_{v_0}$~\cite{kingma2013auto}: a VAE with a 4-layer convolutional encoder and corresponding de-convolutional decoder (same conv-deconv stack is also used for all the other VAE-based methods below). \\
\vspace{1px}
-- $V\!\!AE_{rpl}$: a VAE with a replay buffer that retains 50\% of initial frames from the beginning of training and replays them throughout training. This is our modification of the basic VAE to ensure consistent performance on frames coming from a wider range of RL policies. We included this replay strategy into the rest of the algorithms below, since it helped improve performance in all cases.\\
\vspace{1px}
-- $\beta\text{-}V\!\!AE$~\cite{higgins2017beta}: a VAE with an additional $\beta$ parameter in the variational objective that encourages disentanglement of the latent state. To give $\beta\text{-}V\!\!AE$ its best chance we tried a range of values for $\beta$. \\
\vspace{1px}
-- $SV\!\!AE$: a sequential VAE that is trained to reconstruct a sequence of frames $x_{1},...,x_{T}$ and passes the output of the convolutional stack through LSTM layer before decoding. Reconstructions for this and other sequential versions were also conditioned on actions $a_{1},...,a_{T}$. \\
\vspace{1px}
-- $P\!RED$: a VAE that, given a sequence of frames $x_{1},...,x_{T}$, constructs a predictive sequence $x_{1},...,x_{T\text{+}L}$. First, the convolutional stack is applied to each $x_t$ as before; then, the $T$ output parts are aggregated and passed through fully connected layers. Their output constitutes the predictive latent state. To decode: this state is chunked into $T\text{+}L$ parts, each fed into deconv stack for reconstruction. \\
\vspace{1px}
-- $DS\!A$~\cite{yingzhen2018disentangled}: a sequential autoencoder that uses structured variational inference to encourage separation of static vs dynamic aspects of the latent state. It uses LSTMs in static and dynamic encoders. To give $DS\!A$ its best chance we tried uni- and bidirectional LSTMs, as well as replacing LSTMs with GRUs, RNNs, convolutions and fully connected layers. \\
\vspace{1px}
-- $SP\!AIR$~\cite{SPAIR19}: a spatially invariant and faster version of AIR~\cite{AIR16} that imposes a particular structure on the latent state. $SP\!AIR$ overlays a grid over the image (e.g. 4x4=16, 6x6=36 cells) and learns `location' variables that encode bounding boxes of objects detected in each cell. `Presence' variables indicate object presence in a particular cell. A convolutional backbone first extracts features from the overall image (e.g. 64x64 pixels). These are passed on to further processing to learn `location',`presence' and `appearance' of the objects. The `appearance' is learned by object encoder-decoder, which only sees a smaller region of the image (e.g. 28x28 pixels) with a single (presumed) object. The object decoder also outputs transparency alphas, which allow rendering occlusions.

\textbf{Neural network architectures and training parameters:}

In our experiments, unsupervised approaches learn from 64x64 pixel images, which are rendered by the simulator. 
All approaches (except $SP\!AIR$) first apply a convolutional stack with 4 hidden layers, (with [64,64,128,256] conv filters). The decoder has analogous de-convolutions. Fully-connected and recurrent layers have size 512. Using batch/weight normalization and larger/smaller network depth \& layer sizes did not yield  qualitatively different results. The latent space size is set to be twice the dimensionality of the true low-dimensional state. For VAE we also tried setting it to be the same, but this did not impact results. 
$P\!RED,SV\!AE,DS\!A$ use sequence length 24 for pendulums \& 16 for locomotion (increasing to 32 yields similar results).
$SP\!AIR$  parameters and network sizes are set to match those in~\cite{SPAIR19}. We experimented with several alternatives, but only the cell size had a noticeable effect on the final outcome. We report results for 4x4 and 6x6 cell grids, which did best.

To decouple the number of gradient updates for unsupervised learners from the simulator speed: frames for training are re-sampled from replay buffers. These keep 5K frames and replace a random subset with new observations collected from 64 parallel simulation environments, using the current policy of an RL learner.
Training hyperparameters are the same for all settings (e.g. using Adam optimizer~\cite{kingma2014adam} with learning rate set to $1e\text{-}4$). Since different approaches need different time to perform gradient updates, we equalize resources consumed by each approach by reducing the batch size for the more advanced/expensive learners. $V\!\!AE_{v_0}, V\!\!AE_{rpl}, \beta\text{-}V\!\!AE$ get 1024 frames per batch; for sequential approaches ($SV\!\!AE, P\!RED, DS\!A$) we divide that by the sequence length; for $SP\!AIR$ we use 64 frames per batch (since $SP\!AIR$'s decoding process is significantly more expensive). 

\textbf{Reconstructions for benchmarks and the new multi-object domains:}

\begin{figure}[t]
\centering
\includegraphics[width=0.07\textwidth]{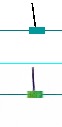}
\includegraphics[width=0.08\textwidth]{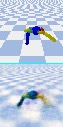}
\vrule
\hspace{1px}
\includegraphics[width=0.07\textwidth]{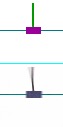}
\includegraphics[width=0.08\textwidth]{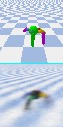}
\vrule
\hspace{1px}
\includegraphics[width=0.07\textwidth]{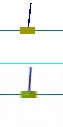}
\includegraphics[width=0.08\textwidth]{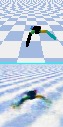}
\vrule
\hspace{1px}
\includegraphics[width=0.07\textwidth]{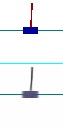}
\includegraphics[width=0.08\textwidth]{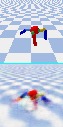}
\vrule
\hspace{1px}
\includegraphics[width=0.15\textwidth]{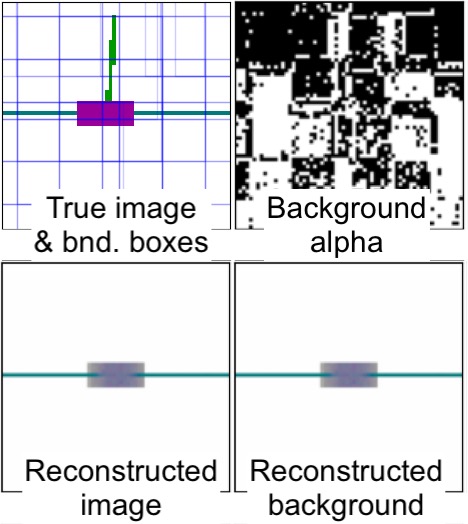}
\includegraphics[width=0.15\textwidth]{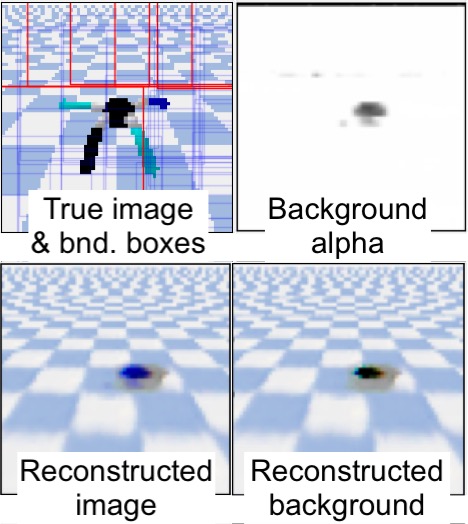}

\hspace{2px}
\hspace{10px}
\includegraphics[width=0.065\textwidth]{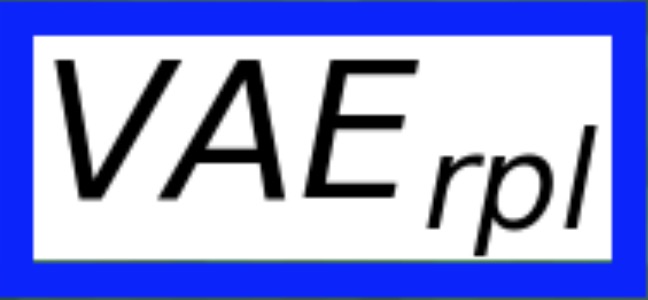}
\hspace{35px}
\includegraphics[width=0.065\textwidth]{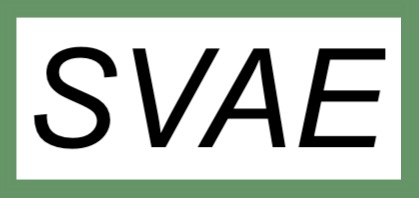}
\hspace{40px}
\includegraphics[width=0.075\textwidth]{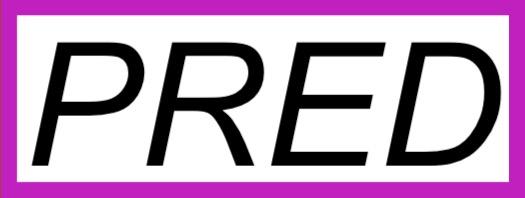}
\hspace{37px}
\includegraphics[width=0.055\textwidth]{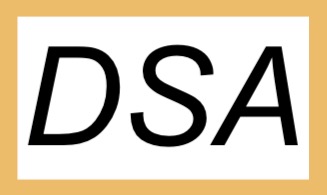}
\hspace{62px}
\includegraphics[width=0.11\textwidth]{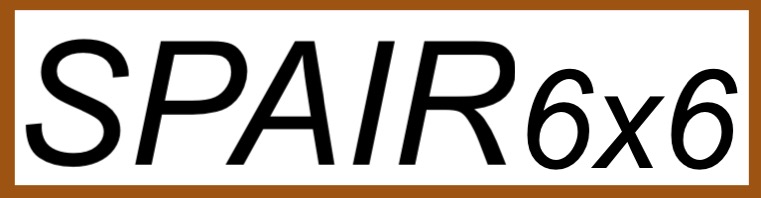}
\hspace{35px}
\vspace{-4px}
\caption{Streaming/unseen frames (top) and reconstructions (bottom) after 500 training epochs.}
\label{fig:bench_recon}
\vspace{-15px}
\end{figure}

Reconstruction for benchmark domains (e.g. \textit{CartPole}, \textit{InvertedPendulum}, \textit{HalfCheetah}, \textit{Ant}) was tractable for $V\!\!AE, SV\!\!AE, P\!RED, DS\!A$. Decoded images were sharp when these algorithms were trained on a static dataset of frames. However, then trained on streaming data with a changing RL policy, decoding was more challenging. It took longer for colors to emerge, especially for $SV\!\!AE$ and $DS\!A$. Sometimes robot links were missing, especially for poses that were seen less frequently.

We attempted to run $SP\!AIR$ on these benchmark domains as well. However, it had difficulties with reconstruction. The thin pole in \textit{CartPole} domain was completely lost, and $SP\!AIR$ mistook the cart base as a part of background. For \textit{HalfCheetah} and \textit{Ant}: a bounding box was detected around the robot, signifying that $SP\!AIR$ did separate it from the background. However, cheetah robot was reconstructed only as a faint thin line, and legs of the \textit{Ant} were frequently missing. 
Right set of plots in Figure~\ref{fig:bench_recon} shows examples of reconstructions, red bounding boxes show detected foreground regions; blue boxes indicate inactive boxes.
$SP\!AIR$ is not specifically designed for domains like this, since its strengths are best seen in identifying/tracking separate objects. Thin object parts and dynamic backgrounds in the benchmark domains are not the best match for $SP\!AIR$'s strongest sides.

\begin{figure}[H]
\vspace{-5px}
\centering
\includegraphics[width=0.1\textwidth]{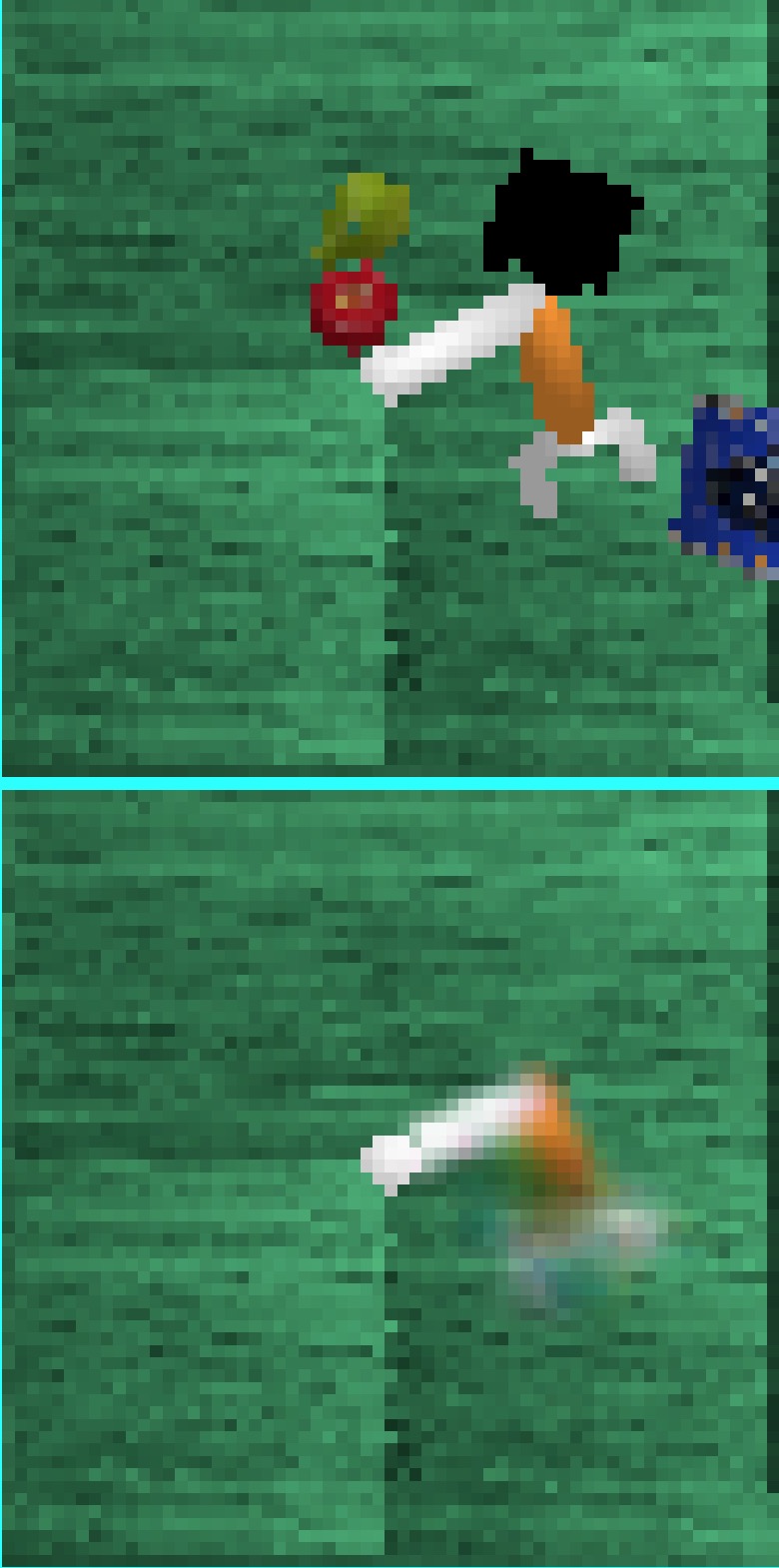}
\includegraphics[width=0.1\textwidth]{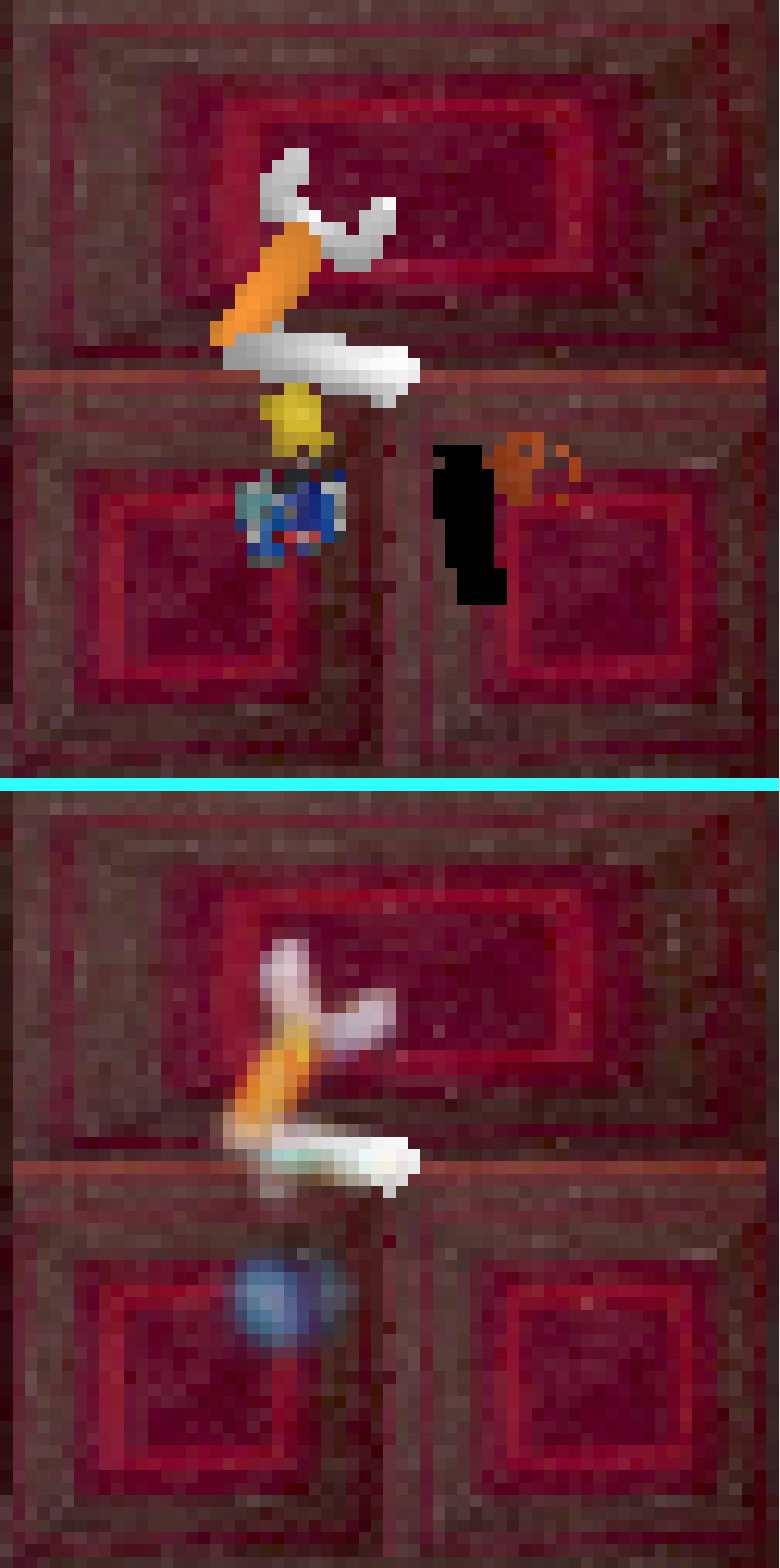}
\includegraphics[width=0.18\textwidth]{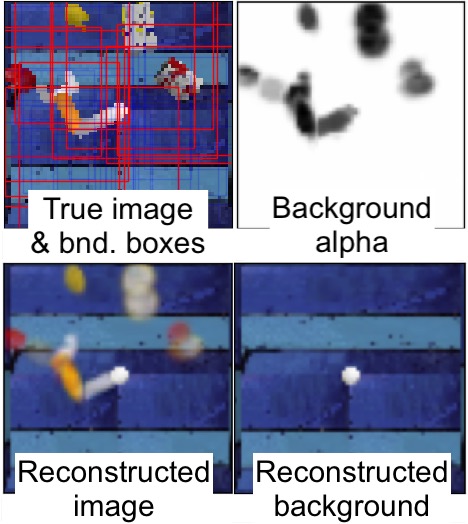}
\includegraphics[width=0.18\textwidth]{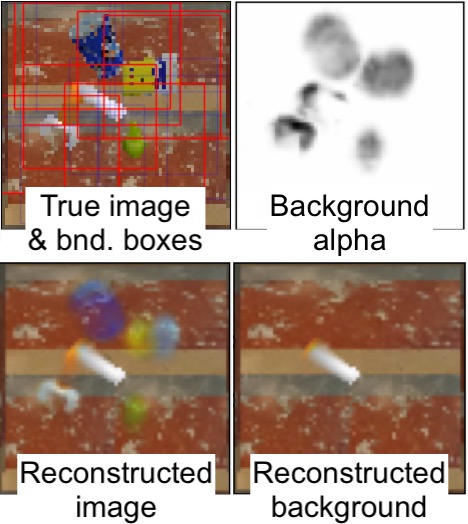}
\vrule width 2pt
\hspace{0.5px}
\includegraphics[width=0.1\textwidth]{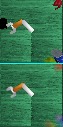}
\includegraphics[width=0.1\textwidth]{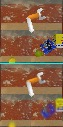}
\includegraphics[width=0.18\textwidth]{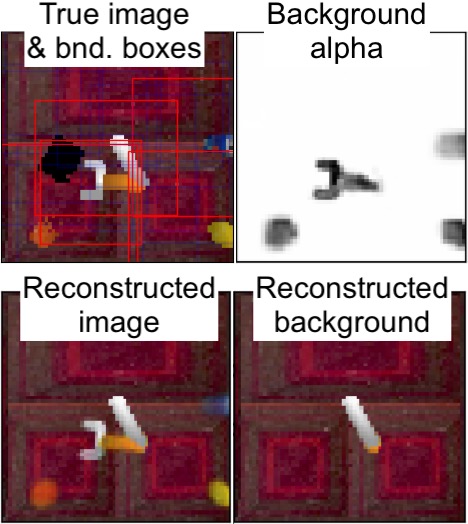}
\includegraphics[width=0.1\textwidth]{img/recon/SPAIR6x6_label}
\includegraphics[width=0.1\textwidth]{img/recon/SPAIR6x6_label}
\hspace{14px}
\includegraphics[width=0.1\textwidth]{img/recon/SPAIR6x6_label}
\hspace{29px}
\includegraphics[width=0.1\textwidth]{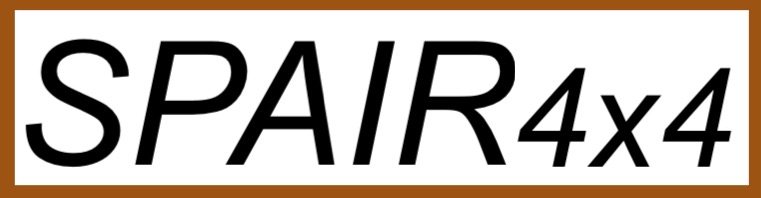}
\hspace{18px}
\includegraphics[width=0.1\textwidth]{img/recon/SPAIR4x4_label}
\includegraphics[width=0.1\textwidth]{img/recon/SPAIR4x4_label}
\hspace{14px}
\includegraphics[width=0.1\textwidth]{img/recon/SPAIR4x4_label}
\hspace{10px}
\vspace{-4px}
\caption{Left side: SPAIR \textit{RearrangeYCB} results after 10K epochs. Right side: SPAIR after 100K epochs. True images are in the top row, reconstructions in the bottom. Thin red bounding boxes overlaid over true images (in the top row) show that bounding boxes did not shrink with further training. SPAIR 6x6 tended to split large objects into  pieces (visible in the case with blue background). SPAIR 4x4 did not split objects and had better results for low-dimensional alignment.
}
\label{fig:spair_recon}
\vspace{-10px}
\end{figure}

As we noted in the main paper, all existing approaches we tried had difficulties decoding \mbox{\textit{RearrangeYCB}} domain. $SP\!AIR$ did manage to produce reasonable reconstructions, albeit missing/splitting of objects was still common.
Figure~\ref{fig:spair_recon} shows example reconstructions after training for 10K epochs ($\approx\!\!32$ hours) and after 100K epochs.
Bounding boxes reported by $SP\!AIR$ were not tight even after 100K epochs (up to 11 days of training overall on one NVIDIA GeForce GTX1080 GPU).
We used PyTorch implementation from~\cite{SPAIRpytorch}, which was tested in~\cite{yonk2020msthesis} to reproduce the original $SP\!AIR$ results (and we added the capability to learn non-trivial backgrounds). An optimized Tensorflow implementation could potentially offer a speedup, but PyTorch has an advantage of being more accessible and convenient for research code.

\begin{figure}[H]
\centering
\vspace{-10px}
\includegraphics[width=0.092\textwidth]{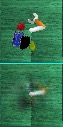}
\includegraphics[width=0.092\textwidth]{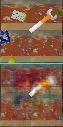}
\includegraphics[width=0.092\textwidth]{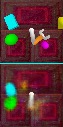}
\includegraphics[width=0.092\textwidth]{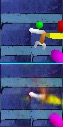}
\vrule
\hspace{1px}
\includegraphics[width=0.092\textwidth]{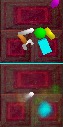}
\includegraphics[width=0.092\textwidth]{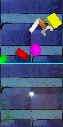}
\vrule
\hspace{1px}
\includegraphics[width=0.092\textwidth]{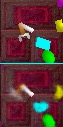}
\includegraphics[width=0.092\textwidth]{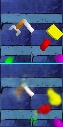}
\vrule
\hspace{1px}
\includegraphics[width=0.092\textwidth]{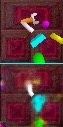}
\includegraphics[width=0.092\textwidth]{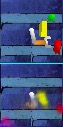}

\hspace{2px}
\includegraphics[width=0.07\textwidth]{img/recon/VAE_rpl_label}
\hspace{6px}
\includegraphics[width=0.07\textwidth]{img/recon/VAE_rpl_label}
\hspace{6px}
\includegraphics[width=0.07\textwidth]{img/recon/VAE_rpl_label}
\hspace{6px}
\includegraphics[width=0.07\textwidth]{img/recon/VAE_rpl_label}
\hspace{11px}
\includegraphics[width=0.07\textwidth]{img/recon/SVAE_label}
\hspace{5px}
\includegraphics[width=0.07\textwidth]{img/recon/SVAE_label}
\hspace{11px}
\includegraphics[width=0.08\textwidth]{img/recon/PRED_label}
\hspace{2px}
\includegraphics[width=0.08\textwidth]{img/recon/PRED_label}
\hspace{12px}
\includegraphics[width=0.052\textwidth]{img/recon/DSA_label}
\hspace{10px}
\includegraphics[width=0.052\textwidth]{img/recon/DSA_label}
\hspace{4px}
\vspace{-4px}
\caption{True images (top) and reconstructed images (bottom) after 10K training epochs.}
\label{fig:vaes_recon}
\vspace{-5px}
\end{figure}

$V\!\!AE, SV\!\!AE, P\!RED, DS\!A$ did not achieve good reconstructions even on \textit{RearrangeGeom} domain. Figure~\ref{fig:vaes_recon} shows example reconstructions. Hence, in the main paper, for analyzing alignment on \textit{RearrangeGeom} domain we chose $V\!\!AE_{rpl}$ and $SP\!AIR$. We focused on these, since $V\!\!AE_{rpl}$ offered speed and simplicity, while $SP\!AIR$ gave better reconstructions.

\section{{\ouralgofullname}}
\setcounter{figure}{0}
\subsection{Proofs and Technical Background for Mathematical Formulation}
\label{sec:suppl_math_proofs}

\textit{Here we present an extended version of Section 3.1 from the main paper. This version contains proofs for all lemmas and theorems, provides relevant technical background from abstract algebra and geometry. We draw analogies with simpler settings from linear algebra to highlight connections with settings that are common in ML literature.}

Let $\R^N$ be the ambient space of all possible latent state sequences $\tau$ (of some fixed length). Let $\M$ be the submanifold of actual state sequences that a dynamical system from one of our domains could generate (under any control policy)\footnote{
In this work, we use the term `manifold' in the sense most commonly used in the machine learning literature, i.e. without assuming strict smoothness conditions. }.
A common view of discovering $\M$ is to learn a mapping that would produce only plausible sequences as output (the `mapping' view).
Alternatively, a submanifold  can be specified by describing all the equations (i.e. relations) that have to hold for the points in the submanifold. Recall an example from linear algebra, where a submanifold is linear, a.k.a. a vector subspace. This submanifold can be represented as an image of some linear map (the `mapping' view), or as null space of some collection of  linear functions, a.k.a. a system of linear equations. The latter is the `relations' view: specifying which relations have to hold for a point to belong to the submanifold.

\subsubsection{Definitions of Independence for Learning Independent Relations}

We are interested in finding relations that are in some sense independent.
One notion of independence is the \textit{functional independence}.
Relations $g_1,..., g_k$ are said to be functionally independent if there is no (non-trivial) function $f: \R^k \!\rar\! \R$ s.t. $f(g_1(\cdot), ... , g_k(\cdot)) = 0$.
However, with such definition, $g(\tau)$ and $h(\tau) g(\tau)$ could be deemed independent\footnote{$f$ can transform $g$s in any way, but does not have direct access to $\tau$, so $h(\tau)$ can not be a `coefficient'.}, even when $h(\cdot) g(\cdot)$ does not provide an additional interesting relation, e.g. $g(\tau)$ vs $\sin(\tau) g(\tau)$. Hence, we need a stricter version of independence. To describe such a version we use the formalism of modules.

A \textit{module} is the generalization of the concept of vector space, where the coefficients lie in a ring instead of a field.
In our case, both  elements of the module and elements of the ring are functions on $\R^N$.
We observe that the set of functions that vanish on $\M$ is closed under the module operations `$+,-$' and multiplication by ring elements, hence it is a (sub-)module.
Recalling the definition of independence for vectors of a vector space, we note that the default notion of independence for elements of a module is analogous.
In this setting, a \textit{syzygy} $\mf^\ddagger$ is a linear combination of relations $g_1,...,g_k$ with coefficients $f_1,...,f_k$ in the ring of functions.
If there is no syzygy $\mf^\ddagger \!=\! \{f_1,..,f_k\}$ s.t. $\sum_{j=0}^{k} f_j g_j$ vanishes, then $g_1,...,g_k$ are independent.

However, for our case the above notion of independence is now too strict, because it would deem any relations $g_1,g_2$ dependent: $g_1 \cdot g_2 - g_2 \cdot g_1=0$ holds for any $g_1,g_2$.
We propose several strategies to avoid this problem.
One option is to define \textit{restricted syzygies}, presented below.
\\


\begin{definition31}[Restricted Syzygy]
Restricted syzygy for relations $g_1,...,g_k$ is a syzygy  with the last entry $f_k$ equal to $-1$, i.e.  
$\mf = \{f_1,..., f_{k-1},f_k\!=\!-1\}$ with $\sum_{j=1}^k f_j g_j \!=\! 0.$ 
\end{definition31}

\begin{definition32}[Restricted Independence]
$g_k$ is independent from $g_1,...,g_{k-1}$ in a restricted sense if the equality $\sum_{j=1}^k f_j g_j \!=\! 0$ implies $f_k\neq-1$, i.e. if there exists no restricted syzygy for $g_1,...,g_k$.
\end{definition32}

For $\mf=\{f_1,..., f_{k-1}, f_k=-1\}$ we denote $\sum_{j=1}^k f_j(\tau) g_j(\tau)$ by $\mf(\tau, g_1, ..., g_k)$.
\vspace{-3px}

Using definitions above, we construct a practical algorithm for learning an (approximately) independent set of relations.
The overall idea is: while learning $g_k$s, we are also looking for restricted syzygies $\mf(\tau, g_1, ..., g_k) \!=\! 0$. Finding them would mean $g_k$s are dependent (in the sense of Definition~3.2), so we augment the loss for learning $g_k$s to push them away from being dependent. We proceed sequentially: first learning $g_1$, then learning $g_2$ while ensuring no restricted syzygies appear for $\{g_1,g_2\}$, then learning $g_3$ and so on.

For training $g_k$s we use \textit{on-manifold} data: $\tau$ sequences come from our dynamical system (i.e. satisfying physical equations of motion, etc). Restricted syzygies $\mf$ are trained using \mbox{\textit{off-manifold}} data: sequences that do not lie on our data submanifold.  We denote such subsequences as 
$\tau_{o\!f\!f}\!=\! \{s_{o\!f\!f_t},s_{o\!f\!f_{t+1}},...,s_{o\!f\!f_T}\}$. Off-manifold data is needed for $\mf$ since we  aim for independence of $g_k$s on $\R^N$, not restricted to their output on data that lies on $\M$ (when restricted to $\M$ the $g_k$s are zero, and so are trivially dependent).   $\tau_{o\!f\!f}$ do not lie on our data submanifold and can come from thickening of on-manifold data or can be random (when $\R^N$ is large, the probability a random sequence satisfies equations of motion is insignificant).

Observe that independence in the sense of Definition~3.2 is the same as saying that $g_k$ does not lie in the \textit{ideal} generated by $(g_1,...,g_{k-1})$, with \textit{ideal} defined as in abstract algebra\footnote{
In the language of abstract algebra: we consider functions on $\R^N$ as module over itself. When a ring is viewed as a module over itself, a submodule of a ring is called an \textit{ideal}. Thus the set of relations that hold on $\M$ is an ideal, called `the ideal of $\M$', written $I(\mathcal{M})$. When considering only subsets of   relations that hold on $\M$, we  will also talk about the `ideal generated by $(g_1,...g_k)$', which is, by definition, the smallest ideal containing $g_1, ..., g_k$.  One can show that this ideal consists of all linear combinations of $g_1, ..., g_k$ with functions as coefficients.
}.
Hence the ideal generated by $(g_1, \ldots,g_{k-1}, g_{k})$ is strictly larger than that generated by $(g_1, \ldots,g_{k-1})$ alone, because we have added at least one new element (the $g_k$). Below we prove that in our setting the process of adding new independent $g_k$s will terminate.
\\

\begin{theorem31}
When using Definition~3.2 for independence and real-analytic functions to approximate $g$s, the process of  starting with a relation $g_1$ and iteratively adding new independent $g_k$s will terminate.
\end{theorem31}
\vspace{-5px}

\begin{proof}
First, we assume that the values of each dimension $d$ of $\tau$ lie between some minimum constants $c_d$ and maximum $C_d$. This is to model actual data observations that are limited by real-world boundaries. This implies that instead of working with unrestricted ambient space, we will work with a compact box $B$, and the corresponding subset of the data manifold $\M_B=\M \cap B$. 
The precise values of $c_d$s, $C_d$s and even the rectangular shape of the box $B$ are immaterial; what is needed is that $B$ is compact and is cut out by a collection of analytic inequalities. In technical terms: we require that $B$ is compact and  \emph{real semi-analytic}.  To avoid boxes with pathological shapes we require in addition that $B$ is the closure of its interior $\mathring{B}$. Possible $B$s include a closed ball, or an arbitrary convex polytope.

We consider the case of using neural networks for approximating relations $g_1, \ldots, g_k$. For networks with real-analytic activation functions (e.g. sigmoid, tanh), the $g$s and relations between them would be real-analytic (recall that a function is analytic if it is locally given by a convergent power series).  
The $g_k$ being independent in the sense of Definition~3.2 implies $g_k$ is not in the ideal generated by $(g_1,...,g_{k-1})$ inside the ring of real-analytic functions.
This means that $(g_1), (g_1, g_2), \ldots, (g_1, \ldots, g_k)$ is a strictly increasing sequence of ideals inside the ring of real-analytic functions on the ambient space $B$. A theorem of J. Frisch  \cite[Th\'{e}or\`{e}me (I, 9)]{frisch1967points}  says that the ring of analytic functions on a compact real semi-analytic space $B$ is \textit{Noetherian}, meaning that any growing chain of ideals in it will stabilize.
This means that after a finite number of iterations we would be unable to learn a new independent $g_k$, meaning we would have found all analytic relations  that hold on $\M_B$, thus terminating the process.
\end{proof}
\vspace{-5px}

If $\M$ itself  is cut out by a finite set of equations of type $h(\tau)\!=\!0$ for some finite set of real-analytic $h$s), then after the process terminates, the subset of $B$ where all relations $g_1,..,g_k$ hold will be precisely $\M_B$. This is the same as saying that all the $h$s defining $\M$ will be in the ideal $(g_1, \ldots, g_k)$. If $\M$ is not cut out by global real-analytic relations, the process will still terminate, having learned all possible global analytic relations that hold on $\M_B$.

We remark that by  a theorem of Akbulut and King \cite{akbulut1992approximating} any smooth submanifold of $\mathbb{R}^N$ can be approximated arbitrarily well by $\M$ defined by a finite set of analytic equations $g(\tau)\!=\!0$.  The same is true even when $g(\tau)$ are restricted to be polynomial. This means that if one ignores the issues of complexity of the defining equations $g$, the differences between various categories of manifolds (smooth, analytic, or algebraic) could be ignored. The above may seem to suggest that methods based on polynomials may suffice. In practice, the  polynomial relations needed may be of very high degree. Hence, using neural networks to learn (approximate) relations would be more suitable.

We further note that  in practice we of course don't have access to $\M$ or even $\M_B$, but only to a finite sample of data points in $\M_B$. The fact that finding independent $g_i$'s vanishing at these points will terminate is a (simpler) special case of the Theorem~3.1, which guarantees that even the more complicated idealized set of relations defining $\M_B$ can be learned in finite time.

Observe that if $g_k$ is dependent on $g_1, \ldots, g_{k-1}$ then the set of points where $g_k$ is zero contains the set of points where all the other $g_1,...,g_{k-1}$ are zero. The converse is not true: while $g_k$ is different from the previous relations in a non-trivial way, it might happen that adding $g_k$  as a relation does not restrict the learned manifold to a smaller set. This arises because of the non-linearity in our setting\footnote{ This is in contrast to linear algebra, where adding an independent linear equation necessarily decreases the dimension of the subspace of solutions.}. 

To ensure that each new relation decreases the data manifold dimension, we could additionally prohibit $g_1,..., g_k$ from having any syzygy $\{f_1,...,f_k\}$ in which $f_k$ itself is not expressible in terms of $g_1,..., g_{k-1}$. This is encoded in the definition below.
\\

\begin{definition33}[Strong Independence]\label{def:strong-independence}
$g_k$ is  strongly independent from $g_1,...,g_{k-1}$  if the equality $-f_k g_k = f_1 \cdot g_1 + ... + f_{k-1} \cdot g_{k-1}$ implies that $f_k$ is expressible as $f_k=h_1 \cdot g_1 + ... + h_{k-1} \cdot g_{k-1}$.
\end{definition33}

In Theorem~3.2 we will show that imposing relations $(g_1, \ldots, g_k)$, such that each new relation is strongly independent from the previous ones, restricts data to a submanifold of codimension at least $k$. 
Since we don't assume that $\M$ has to be smooth, the notion of dimension needs to be defined precisely.
Thus, before embarking on a formal statement and a proof of Theorem~3.2, we give such a definition and discuss related notions needed in the proof.

\subsubsection{Definitions of Dimension in Geometry and Algebra}

For smooth manifolds, which are locally homeomorphic to some $\mathbb{R}^n$, the dimension is simply defined to be $n$, and the invariance of dimension theorem of Brouwer (see \cite[ Theorem 2.26]{hatcher2002algebraic}) ensures that this is unambiguous (which, in light of Cantor's proof that all $\mathbb{R}^n$s have the same number of points and Peano's construction of space-filling curves is not as obvious as it may seem a priori). 

For arbitrary subsets $X$ of $\mathbb{R}^n$ one can then analogously define  $\dim X = d$ if and only if $X$ contains an open set homeomorphic to an open ball in $R^n$, but not an open set homeomorphic to an open ball in $\mathbb{R}^e$, for $e > d$.  We will call this \textit{geometric dimension}. This is the definition we will use when referring to dimension of $\M$.

Now suppose $X$ is a semi-analytic subset of $\mathbb{R}^n$, meaning a subset locally defined by a system of analytic equations and inequalities\footnote{The manifolds we are learning are actually much nicer: they are globally defined by analytic equations. This means, by definition,  that they are $C$-analytic sets (an abbreviation of Cartan real analytic sets; see  \cite[Definition 1.5]{acquistapace2017some} and \cite{cartan1957varietes}, particularly the Paragraphe 11).}.  While $X$ is in general not smooth, it admits a decomposition into smooth parts.  Then, the  definition of geometric dimension $\dim X$ given above coincides with just taking the largest dimension of any part (see, for example, \cite[Proposition 2.10 and Remark 2.12]{bierstone1988semianalytic}). 
This definition is \emph{local}, meaning that if we define dimension of $X$ at a point $\tau\in X$, denoted by $\dim_{\tau} X$, to be dimension of $X\cap U$ for all sufficiently small open neighborhoods $U$ of $\tau$, then $\dim X=\sup_{\tau} (\dim_{\tau} X) $. Of course one also has that $Y\subset X$ implies $\dim Y \leq \dim X$. See \cite[II.1.1]{lojasiewicz1991introduction} for all this and more.

In order to relate this dimension to properties of the relations $g_i$ that define $\M$, we need to connect $\dim \M$ to dimensions of algebraic objects arising from $g_i$s.  These will be rings of various kinds. Thus, we need theory of dimensions of rings.

In commutative algebra the standard way to define a dimension of a ring is due to Krull. It says that dimension of a ring $R$, denoted $\krdim R$, is the length $d$ of the longest chain of prime ideals  $I_d\subsetneq I_{d-1} \subsetneq ... \subsetneq I_{0}$ in $R$.  Note that this has some resemblance to the fact that dimension of a vector space is equal to the length $d$ of the longest chain of subspaces $V=V_d\supsetneq V_{d-1} \supsetneq ... \supsetneq V_0$. For an ideal $I\subset R$ the Krull  dimension is defined as $\krdim I \vcentcolon=\krdim (R/I)$, where $R/I$ is the quotient ring.
See \cite[Chapter 8 and onwards]{eisenbud1995commutative}.

\subsubsection{Statement and Proof of Theorem~3.2}

\begin{theorem32}
Suppose $g_1, \ldots, g_k$ is a  sequence of analytic functions on $B$, each strongly independent of the previous ones. Denote by  $\M_{\mathring{B}}=\{\tau\in \mathring{B}| g_j(\tau)=0 \text{ for all } j\}$ the part of the learned data manifold lying in the interior of $B$. Then dimension of $\M_{\mathring{B}}$ is at most $N-k$.
\end{theorem32}

\textit{Proof outline:}
Strong independence (Definition 3.3) is directly related to the definition of regular sequences.
The proof ultimately aims to use Proposition 18.2 in \cite{eisenbud1995commutative}, which ensures that ideals defined by regular sequences have low dimension. To deduce that $\M$ has low dimension, we need to relate the Krull dimension of the ideal $(g_1, ..., g_k)$ to the geometric dimension of $\M$. To do this we pass through a number of intermediate stages. First we localize, and complexify. This allows us to equate the dimension of the local complexified ideal $(g_1, ..., g_k)$ to that of the local complexified ideal of $\M$, which we do by using local analytic Nullstellensatz. We also equate the  common dimension  of these two ideals to the (local complex) geometric dimension of $\M$. Then, we relate this to local real dimension  of $\M$.  Finally, we get a bound on the (global) dimension of $\M$ itself.
\vspace{-7px}

\begin{proof}
The  Definition~3.3 is equivalent to saying that $g_k$ is not a zero divisor in the ring of functions modulo the ideal generated by $(g_1, \ldots, g_{k-1})$. To see this, we argue as follows.  By definition, in any ring, an element $g$ is not a zero divisor if $fg=0$ implies that $f=0$. Equality  $fg=0$  in the quotient ring means that, in the ring of functions, we have: $fg=\sum_{i=1}^{k-1} f_i g_i$. Thus if $g$ is not a zero divisor in the quotient ring,  then  $fg=\sum_{i=1}^{k-1} f_i g_i$ implies $f$ is zero in the quotient ring, that is to say $f=h_1  g_1 + ... + h_{k-1} g_{k-1}$, for some functions $h_1, ..., h_{k-1}$. 

Thus, a sequence $(g_1, g_2, \ldots, g_{k-1}, g_k)$  where each $g_i$ is strongly independent from the previous ones is a \emph{regular sequence}, see \cite[Sections 10.3, beginning of Section 17]{eisenbud1995commutative}. 

Let $\tau$ be a point in $\M_{\mathring{B}}$. We will denote by $\OO_\tau$  the ring of germs\footnote{A \textit{germ} of a function at a point $\tau$ is an equivalence class of functions defined near $\tau$, where $f_1,f_2$ are considered equivalent if there exists an open neighbourhood of $\tau$ s.t. restrictions of $f_1$ and $f_2$ to that neighborhood coincide.} of real-analytic functions defined near $\tau$, which is isomorphic to the ring of convergent power series centered at $\tau$.  We will denote the complex version of this ring by $\OO_{\tau}^{\C}$. 

The \textit{localization} ring $L_\tau$ of the ring of analytic functions on $\mathring{B}$ at a point $\tau$ is defined as the set of equivalence classes of pairs of analytic functions $(f,h)$ s.t. $h(\tau) \neq 0$, with  the equivalence relation $(f_1,h_1) \sim (f_2,h_2) \iff f_1 \cdot h_2 = f_2 \cdot h_1$.
This is a formal way of introducing fractions $f/h$. 
One also has the \textit{localization map} from the original ring  to the localization ring. It sends $f$ to the equivalence class represented by the pair $(f,1)$, where $1$ is the constant function. In our setting, if one identifies the set of equivalence classes with germs, this map performs a `type conversion' from an analytic function $f$ to its germ at $\tau$. 
In fact, the localized ring $L_\tau$ is a subring of the ring of germs $\OO_{\tau}$. Indeed, a fraction $\frac{f}{h}$ defines an analytic function on some open neighborhood of $\tau$ and the corresponding germ depends only on the equivalence class, thus giving a map $L_\tau \to \OO_\tau$. Clearly the germ is zero only when $f$ is zero, so this map is an injection, and $L_\tau$ is a subring of $\OO_{\tau}$.

However, $L_\tau$ is not all of $\OO_{\tau}$, since not every function analytic at $\tau$ is a ratio of two functions analytic on all of $\mathring{B}$. To remedy this, we consider \textit{completions} of both $L_\tau$ and $\OO_{\tau}$, denoted  $\hat{L}_\tau$ and $\hat{\OO}_{\tau}$ with respect to the maximal ideal of germs vanishing at $\tau$. A completion is perhaps most familiar as a procedure that gives real numbers from rational ones, by means of equivalence classes of Cauchy sequences. In the present situation, a sequence of germs is deemed Cauchy if the difference of any two elements with sufficiently high indexes vanishes to arbitrarily high order (this is known as Krull topology). The completion (of either $L_\tau$ or $\OO_{\tau}$) is then isomorphic to the ring of formal power series centered at $\tau$. Indeed, just taking Cauchy sequences of germs of polynomial functions we get that the completion contains all formal power series centered at $\tau$; and any Cauchy sequence (in either $L_\tau$ or $\OO_{\tau}$) is equivalent to one made up of polynomials, and converges to a formal power series.

We now argue as follows.
Since the localization procedure commutes with taking quotients, and the localization map takes non-zero divisors to non-zero divisors (\cite[Section 15.4]{dummit2004abstract}), we conclude that for each $\tau$ the sequence of germs of $g_1, \ldots, g_k$ is a regular sequence in $L_\tau$. On the other hand, by \cite[Lemma 10.67.5 and Lemma 10.96.2]{stacks-project} (as cited in proof of \cite[Lemma 23.8.1.]{stacks-project}) a sequence is regular in a local ring if and only if it is regular in the completion, so $g_1, ..., g_k$ is regular in $\hat{L}_\tau=\hat{\OO}_\tau$, and so also in $\OO_\tau$.

We claim that the corresponding complexified germs form a regular sequence in $\OO_{\tau}^{\C}$ as well. Indeed, if $f_{j+1} g_{j+1}=\sum_{l=1}^{j}f_l g_l$ on neighborhood $U$ of $\tau$ in $\C^n$, then restricting to $U^{\R}=U\cap\mathbb{R}^n$ and taking real and imaginary parts we see   (denoting by $re(f)$ and $im(f)$ the real and imaginary parts of any complex-valued function $f$) that   on $U^{\R}$ we have $re(f_{j+1}) g_{j+1}=\sum_{l=1}^{j}re(f_l) g_l$ and $im(f_{j+1}) g_{j+1}=\sum_{l=1}^{j}im(f_l) g_l$. Since $g_l$'s form a regular sequence we must have $re(f_{j+1})=\sum_{l=1}^{j}a_l g_l$,  $im(f_{j+1})=\sum_{l=1}^{j}b_l g_l$, so that denoting $c_l=a_l+ib_l$ we have $f_{j+1}=\sum_{l=1}^{j}c_l g_l$ on an open neighborhood of $\tau$ in $\R^N$. Then the same is true on an open neighborhood of $\tau$ in $\C^N$, and so $g_{j+1}$s form a regular sequence in $\OO_{\tau}^{\C}$, as wanted.

Thus the depth of the ideal $I=(g_1, \ldots, g_k)$ in $\OO_{\tau}^{\C}$ (defined as the maximal length of a regular sequence of elements in  $I$ \cite[Section 17.2]{eisenbud1995commutative}) is at least $k$. Moreover, depth of the radical of $J=\sqrt{I}$ is the same (\cite[Corollary 17.8]{eisenbud1995commutative}), and by the local complex-analytic Nullstellensatz (for example, \cite[Theorem 7, Section III.A]{gunning65analytic}) $J$ is the ideal of germs of complex-analytic functions vanishing on the zero-locus of the $g_j$s near $\tau$. By Proposition 18.2 in \cite{eisenbud1995commutative}, codimension of $J$ is is at least $k$ (by definition codimension  it is the supremum of lengths of chains of primes descending from $J$, see [Chapter 9]\cite{eisenbud1995commutative}), so  $\dim J+\codim J\leq n$, and so the (Krull) dimension of $J$ is at most $N-k$.

Local structure theorem for analytic sets implies that this is also the local (complex) geometric dimension of $\M^{\C}$ (see \cite[Proposition 1, IV.4.3]{lojasiewicz1991introduction}  or \cite[Section IIIA]{gunning65analytic}).
Near $\tau$ the real vanishing locus $\M$ is then of real dimension at most $N-k$ ( \cite[Proposition 5]{cartan1957varietes}). Thus $\M$ is of local dimension of at most $N-k$ at all points of $\mathring{B}$, and hence $\dim \M_{\mathring{B}} \leq N-k$  as wanted.
\end{proof}

\subsubsection{Learning Transverse Relations}

It is also possible to define an alternative approach that would provide similar dimensionality reduction guarantees, while also ensuring that the learned relations differ to first order.
To this end we utilize a notion of independence based on \textit{transversality} as follows.

\vspace{7px}
\begin{lemma31}
Dependence as in Definition~3.2 implies  $\nabla\!_{\tau} g_k$ and $\nabla\!_{\tau} g_1,..., \nabla\!_{\tau} g_{k-1}$ are dependent.
\end{lemma31}
\begin{proof}
Suppose $g_1,...,g_k$ are dependent in the sense of Definition~3.2, i.e. $g_k \!=\! f_1 \cdot g_1 \!+\! ... \!+\! f_{k\text{-}1} \cdot g_{k\text{-}1}$. We take gradients w.r.t coordinates of $\R^N$ (the ambient space) and obtain:
\setlength\belowdisplayskip{-5pt}
\begin{align*}
\nabla\!_{\tau} g_k &= \nabla\!_{\tau} [ f_1 \cdot g_1 ] + ... + \nabla\!_{\tau} [ f_{k\text{-}1} \cdot g_{k\text{-}1} ]
= \textstyle\sum_{j=1}^{k\text{-}1} \Big( f_j \nabla\!_{\tau} g_j + g_j \nabla\!_{\tau} f_j \Big)
\end{align*}

Restricting to points in $\M$ and observing that $g_j\!=\!0$ on $\M$, we get $\nabla\!_{\tau} g_k = \sum_{j=1}^{k-1} f_j \nabla\!_{\tau} g_j$. 
\end{proof}

\begin{definition34}[Transversality]
If for all points $\tau^{(i)} \!\in\! \M$ the gradients of $g_1,..,g_k$ at $\tau$, i.e. $\nabla_{\tau} g |_{\tau^{(i)}}$, are linearly independent, we say that $g_k$ is transverse to the previous relations: $g_k \pitchfork g_1,...,g_{k\text{-}1}$.
\end{definition34}

Using transversality, we deem $g_k$ to be independent from $g_1,...,g_{k-1}$ if  the gradients of $g_k$ do not lie in the span of gradients of $g_1,...,g_{k-1}$ anywhere on $\M$. With this, $g_k$ that only differs from previous relations in higher-order terms would still be deemed as `not new'. This stronger notion of independence would be useful for settings where many relations could be discovered, because it is then better to find relations whose first order behavior differs.

This formulation is natural from the perspective of differential geometry. Let $H_{g_j}$ be the hypersurface defined by $g_j$: the set of points where $g_j\!=\!0$. Each hypersurface $H_{g_1},...,H_{g_k}$ contains $\M$. If gradients of $g_k$ are linearly independent from gradients of $g_1,...,g_{k-1}$, then the corresponding hypersurfaces are said to intersect transversely along $\M$.

\vspace{7px}
\begin{lemma32}
For once differentiable $(g_1,..,g_k)$ s.t. $H_{g_j}$s are transverse along their common intersection $H$, this intersection $H$ is a submanifold of $\R^N$ of dimension $N\!-\!k$.
\end{lemma32}
\begin{proof}
Consider the map $G:\mathbb{R}^N \!\to\! \mathbb{R}^k$ given by $G\!=\!(g_1, \ldots, g_k)$.   The fact that $H_{g_j}$s are transverse along $H$ means that the derivative $DG(p)$ has rank $k$ for any $p \in H$. This means that we can pick $k$ linearly independent columns of $DG(p)$. We renumber the coordinates of $\mathbb{R}^N$ so that the ones corresponding to these columns become the first $k$ and apply the Implicit Function Theorem, e.g. 
\cite[Theorem 9.28. p.224]{rudin1976principles}. We can conclude that a neighborhood of $p \in H$ is diffeomorphic to an open set in $\mathbb{R}^{N\!-\!k}$. Since this holds near each $p\in H$, we conclude that $H$ is a manifold of dimension $N\!-\!k$, as wanted\footnote{Our proof is a variation on Preimage Theorem \cite[p. 21]{guillemin1974differential}, and can also be deduced from it:
$H$ is the preimage of  $\vec{0}$ under the map $G$, and  $\vec{0}$  is a regular value because $H\!_{g_j}\!$s are transverse along $H$, implying the lemma. Though note that the Preimage Theorem is itself a direct consequence of the Implicit Function Theorem.}.
\end{proof}
\vspace{-5px}

The notion of independence defined via transversality is infinitesimal and symmetric w.r.t. permuting $g_k$s. This is useful in settings where many relations could be discovered, because it is then better to find relations whose first order behavior differs.
In cases where guaranteed decrease in dimension is not needed, using restricted syzygies could allow a flexible search for more expressive relations.

\subsection{Related Work in Algebraic Ideal Learning}
\label{sec:suppl_math_related_work}

There exists prior work on learning relations carried out in the algebraic setting. 
Some of this work aims to find simple polynomial relations that hold on the data manifold.
The criterion for simplicity is the polynomial degree. 
Most of these works find either all relations or all relations of a given degree at the same time. This is in contrast to our approach, which finds relations one by one, making it amenable to finding as many relations as desired.
Moreover, since we aim to use neural networks for learning relations, the class of polynomial relations is not suitable for our purposes. Hence, we consider a substantially different setting of learning analytic relations. The notion of degree is not defined for analytic functions, making work based on this notion not directly applicable to our setting. In contrast to the algebraic case, our notion of simplicity is implicit in the expressivity of the networks.  However, some of the issues that arise in our setting have parallels in the algebraic setting. Below we give a brief overview, pointing out these connections.

The problem of learning relations that hold approximately on a given dataset was brought to the  machine learning community by~\cite{livni2013vanishing}.
This paper introduced the algorithm called Vanishing Component Analysis (VCA). The VCA algorithm depends on a parameter $\varepsilon$, and in the limit $\varepsilon=0$ finds a set of generators for the ideal of polynomials that vanish on a data set.
The algorithm builds up this set of generators degree by degree, starting with linear ones (if such exist). For general $\varepsilon$ it finds polynomials $P_i$ such that the (Euclidean) norm of the vector of values of $P_i$ is at most $\varepsilon$. The specifics of which of these polynomials it finds depend on the inner workings of the algorithm, which is based on SVD. Being a linear algebra based algorithm, it finds all of those polynomials of the specific degree at the same time. 

The same problem of learning relations that hold approximately on a given dataset has been considered before in mathematics literature. \cite{sauer2007approximate} introduced $p$-approximate ideals of
accuracy $\varepsilon$, and \cite{heldt2009approximate} introduced $\varepsilon$-approximate vanishing ideals; these are two related but different objects aiming at capturing such approximate relations. One of the differences between them is how they normalize the polynomials. The issue at hand is that if one considers only values of a function $P$ on the data set, then a sufficiently small rescaling of \textit{any}  $P$ will have values that are small, and so will be deemed an approximate relation. Such a rescaled function will have small values in a lot of places, not just near the data set itself, and so would be a `trivial' or `spurious' relation.   To avoid this problem of `spurious approximate relations' one needs to normalize $g$ itself. \cite{sauer2007approximate} considers $L_p$ norms of the coefficient vector (hence the $p$ in the name), while \cite{heldt2009approximate} considers only the $L_2$ norm. In~\cite{kera2019spurious}, it is observed that the VCA algorithm from \cite{livni2013vanishing} does produce such `spurious' small-coefficient polynomials. The authors introduce a modification to VCA in which the values of $f$ on the data set are traded off against its norm, like in \cite{sauer2007approximate} and \cite{heldt2009approximate}. By default \cite{kera2019spurious} also uses   the Euclidean norm of the vector of coefficients of $P$, or some modification (such as truncation) of it. 
In a follow up paper \cite{kera2019gradient}, the `norm' is now  given by (the norms of) the gradient vectors of $P$ on the data points. 

In an alternative formulation of approximate vanishing, which is geometric and avoids the spurious relation problem: one looks for relations whose vanishing loci pass near the data points (rather than the ones which take small values exactly at the data). This approach is more challenging for the algebraic methods but has been attempted in ~\cite{fassino2010almost, fassino2013simple}. We note that it is in fact related to gradient normalization, and this relation underlies a part of our approach.

Observe that, while the setting of our work is very different, the need to decide which relations `hold approximately' on a data set in the presence of rescalings is common to both settings. The norm of the coefficient vector is obviously unavailable in our setting. On the other hand, we employ a transversality framework for multiple relations, which places an emphasis on the on-manifold gradients as its core principle. As a special case, this produces the `singleton-transversality' approach: comparing on-manifold values of $g$ to the on-manifold gradients of $g$ (similar to one used in \cite{kera2019gradient} except that in our case it is formulated as a component of NN loss). More precisely, we use the ratio of the absolute value of $g$ $\big($i.e.$|g(\tau)|\big)$ to the norm of the gradient of $g$ at a data point $\tau$ $\big($i.e.$\norm{\nabla \!\!_{\tau}\!(g)|_{\tau}}\big)$:
$d_g(\tau)=\frac{|g(\tau)|}{\norm{\nabla \!\!_{\tau}\!(g)|_{\tau}}}$. This has the following interpretation: the norm of the gradient measures the maximal rate of change of the linearization of $g$ at $\tau$, meaning the maximal `slope'.
So $d_g(\tau)$ is the distance from $\tau$ to the nearest point where this linearization vanishes ($d_g(\tau)\!=$~height/slope~$=$~distance). This serves as a proxy for the distance from $\tau$ to the vanishing locus of $g$ itself. In this way, our approach unifies the gradient based and distance-to-vanishing-set based measures of approximate vanishing that have appeared in the prior work cited above.
In addition to gradient measures, we also assess the vanishing of $g$ by comparing values on-manifold  and off-manifold, which is related to gradients in spirit (gradients tell you how much the value nearby differs from the values at the point you start with), 
but requires only evaluations of the relation itself. In our case we use this comparison to formulate a stopping criterion for learning relations: stopping when on-manifold values are sufficiently smaller than off-manifold ones.

Learning of algebraic manifolds has been considered in learning theory works, e.g. \cite{hazan2016non, globerson2017effective}. It would be interesting to investigate whether analytic manifolds that we consider, which are less rigid than algebraic ones (but more rigid than smooth, as illustrated by Theorem~3.1, for example), give another reasonable alternative. 

On the applications side,~\cite{iraji2017principal} search for a low-dimensional manifold (variety) cut out by polynomials of bounded degree, and show a proof of concept for data modeling in a robotics setting. VCA algorithm has been applied to pattern recognition by several works, e.g.~\cite{yan2018deep, zhao2014hand}.

\subsection{Additional Details, Results and Illustrations for Evaluating {\ouralgo}}

\subsubsection{Further Algorithmic and Implementation Details}
\label{sec:suppl_b31}

Here, we start by giving a further mathematical interpretation for our implementation of {\ouralgo} with transversality, i.e. the more detailed motivation for Equation~2 in the main paper. Then, we give a summary of implementation details for {\ouralgo}.

\textbf{Motivation for our approach to computing transversality:}

Recall that, to obtain $g_k$ that is transverse to $g_1,..,g_{k-1}$ (Definition~3.4), we compute gradients of each $g_1,...g_{k-1}$ w.r.t the input. For example, for $g_1$ we denote this as $v_1 = \nabla\!_{\tau}(g_1)|_{\tau}$.
Making $g_k$ transverse to $g_1,...g_{k-1}$ means ensuring that $v_k$ is linearly independent of $v_1,...,v_{k-1}$.
Hence, we need to choose a (computationally tractable) numerical measure of this linear independence.
To that end, we design our measure to maximize the angles between $v_k$ and all the previous $v_1,..,v_{k-1}$. When the number of relations is lower than the dimensionality of the ambient space ($k \!\leq\! N$), this is maximized by any vector that is perpendicular to all the previous ones. 
Such a measure also encourages transversality of subsets of relations.
Furthermore, we want to discourage small angles, which can be achieved by a measure that involves a product of pairwise measures.

Hence, we use the product of sines of pairwise angles as our measure of transversality (with log for computational stability):
\begin{align}
L_{tr}(g_k) = d_{g_k}(\tau) - \log\norm{v_k} - \log\textstyle\prod_{j=1}^{k-1} \sin^2(\theta_{v_j,v_k})
\tag{2}
\end{align}
The last two terms give (up to weighting constants) the log of products of areas of parallelograms formed by $v_k$ and each of the previous $v_1,...,v_{k-1}$.
In principle, the $k$-dimensional volume of the parallelepiped spanned jointly by $v_1,...,v_k$ could serve as a measure of transversality. It could be computed as a product of singular values of the matrix with columns $v_1,...,v_k$, e.g. requiring SVD. However, it would not be suitable for low-dimensional cases $(N\!<\!k)$, since this volume would be 0.

\textbf{{\ouralgo} implementation details:}

We implemented {\ouralgo} in PyTorch~\cite{pytorch19}. To represent $g_k$ relations and restricted syzygies $\mf$ we used fully connected networks with 3 hidden layers. For our experiments we used a setup that starts with small networks (e.g. 3 layers, 4 hidden units per layer for $g_1$) and doubles the number of hidden units for subsequent $g_k$s (e.g. 8 hidden units per layer for $g_2$, 16 for $g_3$, and so on, with a maximum of 256). For syzygies we started with 32 nodes per layer. We also experimented with simply having 32 units in all $g_k$s, but did not see a significant difference.

The first term in Equations~1,2,3 in the main paper dictates whether $g_k$ is close to 0 for on-manifold data. Since there are no further weighting terms in these equations, we note that one needs to take care that the other terms do not overwhelm the contribution from the 1st term. For this, we clip the loss from other components if it is more than twice the magnitude of the 1st term. This simply means: the overall loss encourages $g_k$ outputs to be small for on-manifold data, regardless of what other parts dictate.

When learning with transversality: we usually used a fixed weight $\beta\!=\!1e3$ for the transversality terms instead of loss clipping ($\beta \in [1e2,1e4]$ worked as well). When using the variant with restricted syzygies: we always included the second term from Equation~3, i.e $\!\nabla\!_{g_k} \big[ |\mf(\tau_{o\!f\!f},g_1,...,g_k)| \big]$, even if $\mf$ did not reach output close to zero during its training. This implies that we implemented a soft (incremental) version, rather than mandating syzygies to be always learned exactly.

In theory, $g_k$ being 0 for on-manifold data means getting an output of exactly 0. But in practice we need to choose a way to tell whether the output of $g_k$ or $\mf$ is essentially 0 for all practical purposes. So, as our stopping criterion, we look at the difference between on-manifold and off-manifold outputs. When the mean absolute value of off-manifold values is more than 5 times that of on-manifold values: we say we are done learning $g_k$ (or $\mf$). We record the mean outputs for on- and off-manifold data when we save the learned relations. With that, when {\ouralgo} relations are loaded for subsequent use, we can check if the output of $g_k$ is `close to 0': simply check whether it is close to the expected on-manifold output magnitude. To make this more concrete, below is an example of such expected values, printed when a learned set of relations is loaded:

\begin{verbatim}
20:52:07 AML loaded 2 relations, 0 syzygies, on/off means:
20:52:07 [0.0047 0.0029]
20:52:07 [0.0518 0.025 ]
\end{verbatim}

Top row shows mean expected on-manifold output for $g_1,g_2$. Bottom row shows expected off-manifold output means. Note that off-manifold values are $\approx\!\!10$ times higher than on-manifold ones. The relative magnitude is what matters, not whether the values are `small' in some absolute sense.

\subsubsection{Distortion Measure}
\label{sec:suppl_b32}

We use the following measure of distortion of a map $f$: take pairs of inputs $\tau_1$, $\tau_2$ and the corresponding outputs $f(\tau_1), f(\tau_2)$, then compute distortion coefficient $\rho_{distort}$:
\begin{align*}
\rho_{distort}(f)|_{\tau_1,\tau_2} \!=\! \log
\frac{
d_{L2}\big(f(\tau_1), f(\tau_2)\big) }{ d_{L2}(\tau_1, \tau_2)}.
\end{align*}
Here, $d_{L2}$ is the Euclidean distance. A map that yields low variance of these coefficients would better preserve the geometry of the domain (up to overall scale). This measure is related to approaches surveyed in~\cite{distort18, bartal2019dimensionality} and has the same desirable properties as $\sigma$-distortion described in~\cite{distort18}, but in log space.
Observe that, if $h$ is a composition $h = f \circ g$, then
\begin{align*}
\rho_{distort}(h)|_{\tau_1,\tau_2} = \rho_{distort}(g)|_{\tau_1,\tau_2} + \rho_{distort}(f)|_{g(\tau_1),g(\tau_2)}.
\end{align*}
This additivity of individual $\rho$s is appealing. It makes the variance measure defined from them extendable to a distortion covariance measure for composable maps (with inverse maps maximally anti-correlated). This measure is also related to Hilbert distance on rays, which appears in the work of Birkhoff on Perron-Frobenius theory \cite{birkhoff1957extensions}. We plan to further investigate this in future work.

\vspace{7px}
\subsubsection{Additional Illustrations of {\ouralgo} Results}
\label{sec:suppl_b33}

\begin{figure}[t]
\centering
\includegraphics[width=0.23\textwidth]{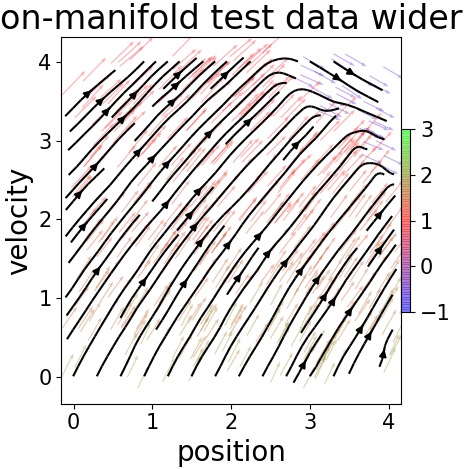}
\includegraphics[width=0.17\textwidth]{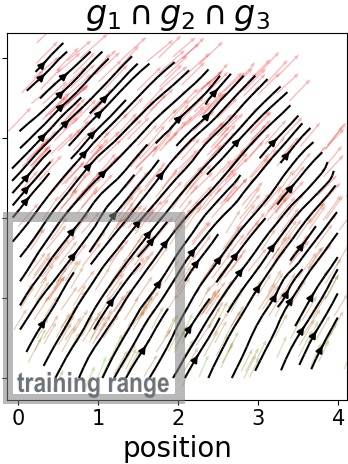}
\includegraphics[width=0.17\textwidth]{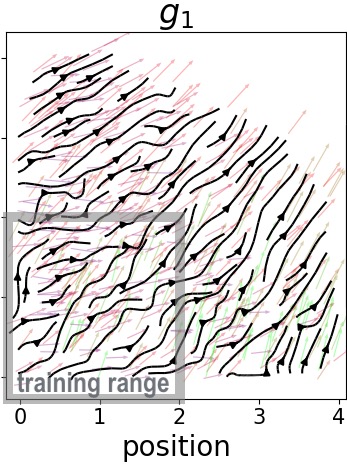}
\includegraphics[width=0.17\textwidth]{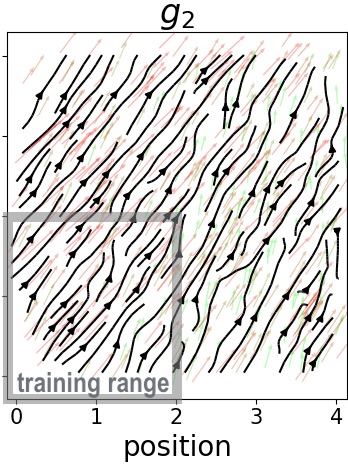}
\includegraphics[width=0.17\textwidth]{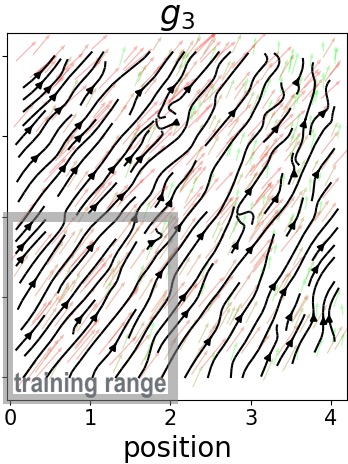}
\\
\includegraphics[width=0.215\textwidth]{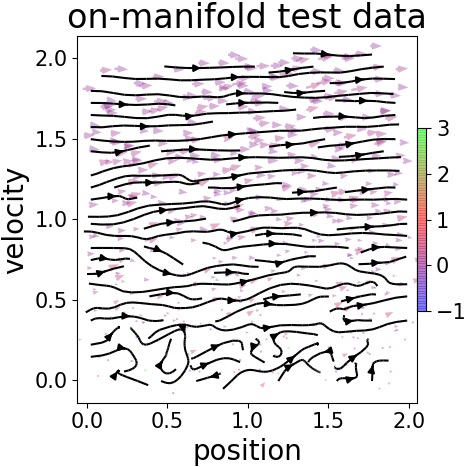}
\includegraphics[width=0.16\textwidth]{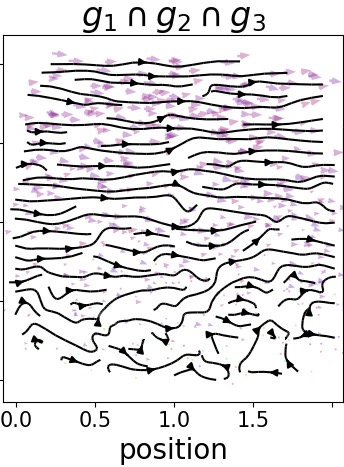}
\includegraphics[width=0.16\textwidth]{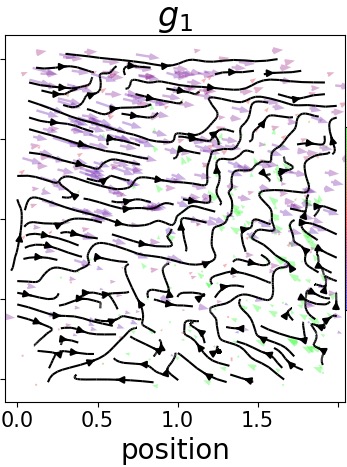}
\includegraphics[width=0.16\textwidth]{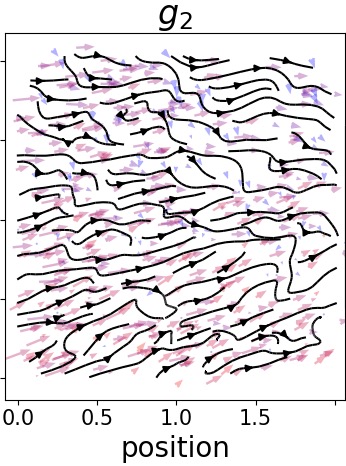}
\includegraphics[width=0.16\textwidth]{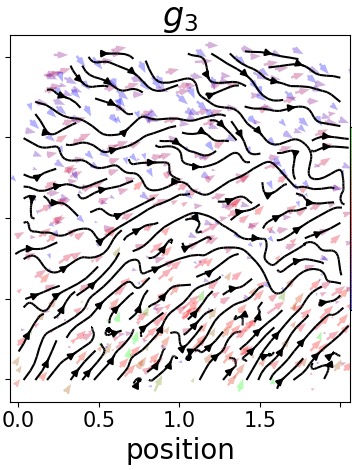}
\\
\includegraphics[width=0.215\textwidth]{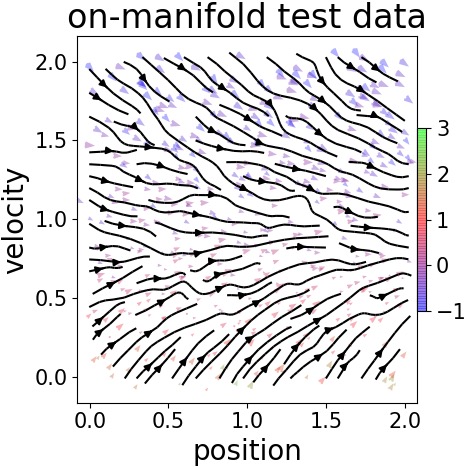}
\includegraphics[width=0.16\textwidth]{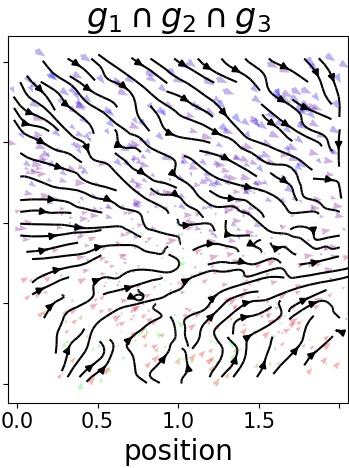}
\includegraphics[width=0.16\textwidth]{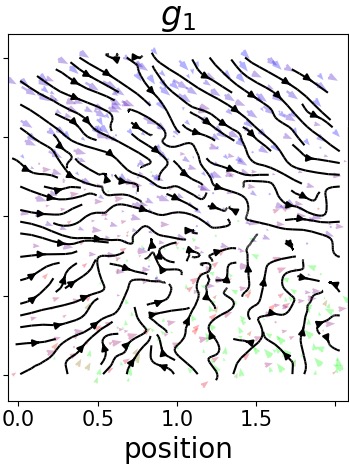}
\includegraphics[width=0.16\textwidth]{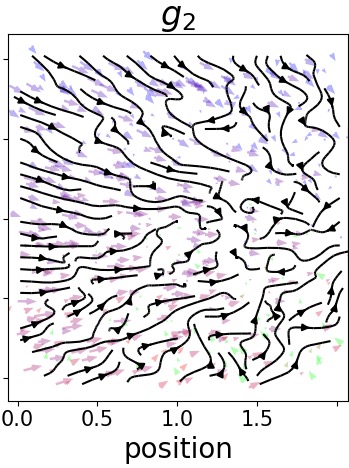}
\includegraphics[width=0.16\textwidth]{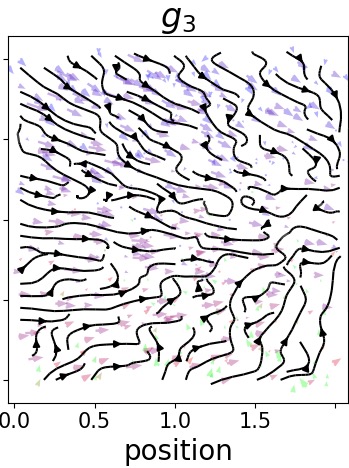}
\vspace{-3px}
\caption{Illustrations of learning relations on the block-on-incline domain with syzygies.}
\label{fig:aml_block_syzygies}
\vspace{-10px}
\end{figure}

Here, we provide additional illustrations of learning relations with {\ouralgo} in the \textit{block-on-incline} domain (Figure~\ref{fig:block_on_incline} in the main paper).
On-manifold data is comprised of noisy position and velocity observations from simulation of a block with mass 1kg sliding down an incline.
Figure~\ref{fig:aml_block_transverse} in the main paper illustrated learning with transversality. Figure~\ref{fig:aml_block_syzygies} (on the next page) shows the corresponding results when using syzygies.
The true dynamics and the learned relations are visualized using phase space plots: arrows indicate change in position \& velocity after $1sec$ of sliding (scaled to fit).

Each row in Figure~\ref{fig:aml_block_syzygies} shows: on-manifold data visualizing the actual dynamics; part of the space where the intersection $g_1 \cap g_2 \cap g_3$ of the learned relations holds (i.e. all $g_1,...,g_k$ output values close to 0); individual preimages of 0 for each relation separately. The top row shows training on a limited range when a block slides on a $45^{\circ}$ incline.
The intersection $g_1 \cap g_2 \cap g_3$ generalizes far beyond the training data ranges. It misses only the part capturing stopping at the end of the incline (blue arrows in top right of `on-manifold test data' plot), which is not possible to extrapolate, since training does not contain examples of running into the end of the incline.
The middle row shows results for a $35^{\circ}$ incline with high friction coefficient $\mu_k\!=\!0.8$. The bottom row shows results when using a high drag coefficient $\mu_d\!=\!2.0$; in this case we train on a range of incline angles $\theta \in [\tfrac{\pi}{20}, \tfrac{\pi}{2.5}]$ and visualize results for a $10^{\circ}$ incline.

Overall, both training with transversality and with syzygies gives us the ability to generalize and capture non-trivial dynamics.
We can see that intersections $g_1 \cap g_2 \cap g_3$ look very similar to on-manifold phase space plots, which means {\ouralgo} correctly captures information about the data manifold. As expected, zero-level sets of individual images do not resemble on-manifold plots, since individual relations $g_k$ capture different parts/aspects of on-manifold data properties.

\vfill

\pagebreak
\bibliographystyle{IEEEtran}
\bibliography{references}
\end{document}